\RequirePackage{fix-cm}
\PassOptionsToPackage{sort&compress}{natbib}
\documentclass[10pt,a4paper,numbers]{nus-techreport}
\usepackage{nus-techreport}
\pdfoutput=1
\usepackage{tikz,needspace,wrapfig,url,placeins,setspace}
\usetikzlibrary{arrows.meta,positioning}

\definecolor{wblue}{RGB}{52,204,204}
\definecolor{wyellow}{RGB}{255,192,0}
\definecolor{wfg}{HTML}{1C2B33}
\usepackage[T1]{fontenc}
\usepackage{bm,multirow,makecell,colortbl,threeparttable}
\usepackage{placeins,enumitem,microtype}
\usepackage{adjustbox}
\newcommand{\modelname}{EponaX}
\DeclareRobustCommand{\model}{%
  \texorpdfstring{%
    \raisebox{\dimexpr-\fontcharht\font`E*34/100\relax}{%
      \includegraphics[height=\dimexpr\fontcharht\font`E*134/100\relax]{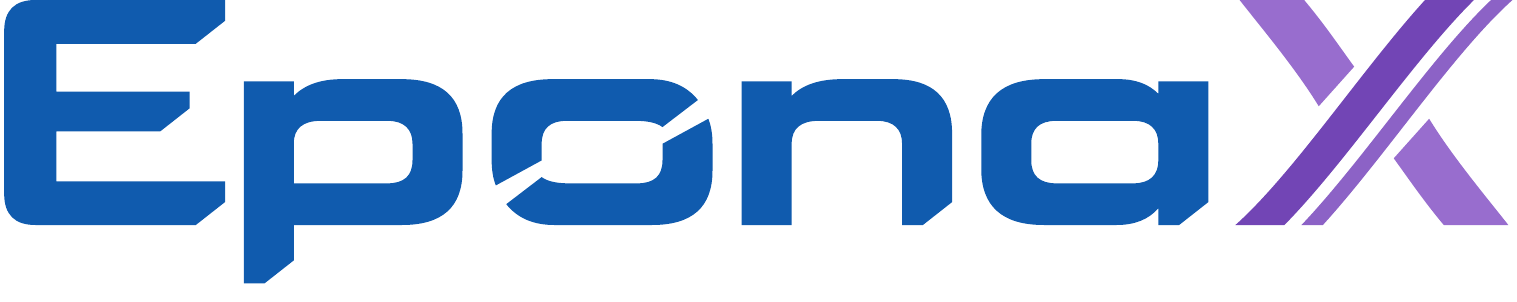}}%
  }{\modelname}%
}

\usepackage{pifont}
\newcommand{\cmark}{\ding{51}}
\newcommand{\xmark}{\ding{55}}

\DeclareMathVersion{navsim}
\DeclareMathSizes{7.7}{7.7}{5}{5}
\SetSymbolFont{operators}{navsim}{T1}{ptm}{m}{n}
\SetMathAlphabet{\mathbf}{navsim}{T1}{ptm}{b}{n}

\definecolor{navsimScoreBlue}{HTML}{2A63AC}
\definecolor{navsimScorePurple}{HTML}{734FA9}
\ExplSyntaxOn
\tl_new:N \l__navsim_score_hue_tl
\tl_new:N \l__navsim_score_text_tl
\fp_new:N \l__navsim_score_value_fp
\fp_new:N \l__navsim_score_min_fp
\fp_new:N \l__navsim_score_max_fp
\fp_new:N \l__navsim_score_position_fp

\cs_new_protected:Npn \__navsim_read_score:n #1
  {
    \tl_set:Nx \l__navsim_score_text_tl { \tl_to_str:n {#1} }
    \regex_replace_all:nnN { [^0-9.] } {} \l__navsim_score_text_tl
    \fp_set:Nn \l__navsim_score_value_fp { \l__navsim_score_text_tl }
  }

\NewDocumentCommand \NavsimPrepareScores { m m m }
  {
    \tl_set:Nn \l__navsim_score_hue_tl {#1}
    \fp_set:Nn \l__navsim_score_min_fp { inf }
    \fp_set:Nn \l__navsim_score_max_fp { -inf }
    \cs_set_eq:NN \__navsim_saved_midrule \midrule
    \cs_set_eq:NN \midrule \prg_do_nothing:
    \cs_set:Npn #3 ##1##2##3##4
      {
        \tl_if_eq:nnF {##1} {Human}
          {
            \__navsim_read_score:n {##4}
            \fp_set:Nn \l__navsim_score_min_fp
              { min(\l__navsim_score_min_fp, \l__navsim_score_value_fp) }
            \fp_set:Nn \l__navsim_score_max_fp
              { max(\l__navsim_score_max_fp, \l__navsim_score_value_fp) }
          }
      }
    #2
    \cs_set_eq:NN \midrule \__navsim_saved_midrule
  }

\NewDocumentCommand \NavsimShadeScore { m m }
  {
    \tl_if_eq:nnTF {#1} {Human}
      { \cellcolor{black!10} }
      {
        \__navsim_read_score:n {#2}
        \fp_set:Nn \l__navsim_score_position_fp
          {
            (\l__navsim_score_value_fp - \l__navsim_score_min_fp) /
            max(0.000001, \l__navsim_score_max_fp - \l__navsim_score_min_fp)
          }
        \use:x
          {
            \exp_not:N \cellcolor
              { \l__navsim_score_hue_tl ! \fp_eval:n { round(8 + 54 * \l__navsim_score_position_fp ^ 1.6, 1) } }
          }
      }
  }
\ExplSyntaxOff

\reportsetup{accent=005AFF,secondary=005AFF,title-size=23pt,title-leading=25pt}
\title{\texorpdfstring{%
  \model:~Recipes for Training and Deploying World-Action Models in the Real World%
}{\modelname: Recipes for Training and Deploying World-Action Models in the Real World}}

\tcbset{eponaxPanel/.style={bicolor,boxsep=0pt,boxrule=0.55pt,arc=1.5mm,
  colframe=reportAccent!24!white,colback=reportAccent!7!white,
  colbacklower=reportAccent!1.5!white,coltext=reportInk,
  left=3mm,right=3mm,top=1.4mm,bottom=1.7mm,middle=1.3mm,
  before skip=0pt,after skip=0pt,
  segmentation style={solid,draw=reportAccent!24!white,line width=0.45pt},
  overlay unbroken={\draw[reportAccent!16!white,line width=0.4pt]
    ([yshift=-1.3mm]segmentation.center) -- ([yshift=1.7mm]interior.south);}}}
\reportmetadata{%
  \sffamily
  {\normalsize\bfseries Horizon Robotics World Model Team}%
  \hspace{1.5em}%
  \href{https://garlicba.github.io/EponaX}{%
    \faGlobe\enspace\textbf{Project Page}\enspace garlicba.github.io/EponaX}}

\newcommand{\eponaxInstitutionLogo}{%
  \includegraphics[trim=69bp 112bp 69bp 113bp,clip,height=11mm]{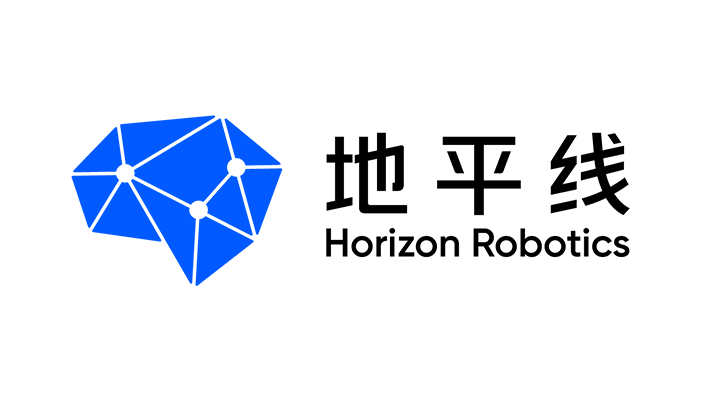}}
\makeatletter
\renewcommand\nr@pdfauthors{Mingxiao Li, Zhijian Shu, Wei Yin, Jiawei Xu,
  Mingkai Jia, Alan Liang, Zhizhou Zhong, Shiyao Li, Dongyue Lu, Liang Yue,
  Zeyuan Xu, Kaicheng Zhang, Xiao-Xiao Long, Lingdong Kong, Qian Zhang}
\preto\nr@maketitle{\noindent\eponaxInstitutionLogo\par\vspace{-3pt}}
\makeatother

\DeclareCaptionFont{vitalCaption}{\fontsize{11pt}{13pt}\selectfont}
\makeatletter
\AtEndCaption{\ifdefstring{\@captype}{table}{\vspace{-0.1cm}}{}}
\makeatother

\DeclareCaptionFont{configurationCaption}{\fontsize{10pt}{12pt}\selectfont}
\newenvironment{trainingconfiguration}[2]{%
  \begin{table}[!htp]
  \setlength{\intextsep}{9pt plus 1pt minus 1pt}
  \begin{tcolorbox}[eponaxPanel]
  \captionsetup{font=configurationCaption,skip=3pt}
  \caption{#1}\label{#2}
  \tcblower
  \small\setstretch{1}\noindent\ignorespaces
}{%
  \end{tcolorbox}
  \end{table}
}
\newenvironment{configurationgroup}[1]{%
  \begin{minipage}[t]{0.477\linewidth}
  \raggedright
  \setlength{\parskip}{0.8pt}
  \noindent\reporthead{#1}\par
  \vspace{1pt}%
}{%
  \end{minipage}%
}
\newcommand{\configurationfield}[2]{%
  \hangindent=66pt\relax
  \noindent\makebox[66pt][l]{\textbf{#1}}#2\par
}
\newcommand{\eponaxPanelNextRow}{%
  \par\vspace{4pt}\noindent
  \textcolor{reportAccent!18!white}{\rule{\linewidth}{0.45pt}}%
  \par\vspace{3pt}\noindent\ignorespaces
}

\BeforeBeginEnvironment{wrapfigure}{\Needspace{19\baselineskip}}
\pretocmd{\section}{\Needspace{6\baselineskip}}{}%
  {\PackageError{vital}{Cannot reserve space before sections}{Check the report class.}}


\newsavebox{\vitalpagebox}
\makeatletter
\renewcommand\ps@nusreport{%
  \def\@oddfoot{%
    \begingroup
    \sffamily\bfseries
    \fontsize{\nr@pagesize}{\nr@pageleading}\selectfont
    \color{reportAccent}%
    \sbox{\vitalpagebox}{\thepage}%
    \ifnum\value{page}>1\relax
      \raisebox{\dimexpr(\ht\vitalpagebox-\dp\vitalpagebox-2.5pt)/2\relax}{%
        \rule{12mm}{2.5pt}}%
    \fi
    \hfil\usebox{\vitalpagebox}%
    \endgroup}%
  \let\@evenfoot\@oddfoot
  \let\@oddhead\@empty
  \let\@evenhead\@empty}
\patchcmd{\reportabstract}{\nr@abstract\par}{%
  \fontsize{11pt}{14.2pt}\selectfont\nr@abstract\par
}{}{\PackageError{vital}{Cannot set abstract body size}{Check the report abstract layout.}}
\makeatother

\tcbset{nrAbstract/.append style={top=3mm,bottom=3mm}}

\abstract{
World-action models promise to turn predictive scene representations directly into driving decisions. Deploying these models on real vehicles requires processing multi-camera context and long-horizon dynamics within the onboard computation budget. The planner must also retrieve the right evidence as its trajectory hypothesis evolves. We introduce \model, which scales world-action modeling by rethinking how driving evidence is represented and converted into action. Our approach preserves fine-grained visual evidence, compresses calibrated surround observations into a compact multi-view representation, and jointly models these visual features with semantically aligned structured navigation. Predicting future appearance, geometry, and semantics during training encourages the shared backbone to encode long-horizon scene dynamics, while deployment retains only the trajectory pathway. For action generation, \model~couples a compact trajectory latent with the complete backbone hierarchy, while TrajLoop updates block-local conditions as the trajectory hypothesis evolves. Trained on approximately 35K hours of real-world driving data, \model~achieves 91.9 PDMS on NAVSIM v1 before GRPO and 92.5 PDMS after GRPO. On NAVSIM v2, it achieves 91.8 EPDMS on navtest after GRPO and a combined 48.3 EPDMS on navhard. \model~sustains onboard planning at \textbf{10\,Hz} over approximately \textbf{240\,km} across four urban routes, averaging \textbf{3--5} safety-related driver interventions per \textbf{100\,km}.
}

\begin{document}
\newgeometry{top=1cm,bottom=0.65cm,left=2.5cm,right=2.5cm}
\let\cite\citep
\maketitle
\vspace{8pt}
\noindent\begin{minipage}{\linewidth}
    \centering
    \captionsetup{hypcap=false}
    \includegraphics[width=\linewidth]{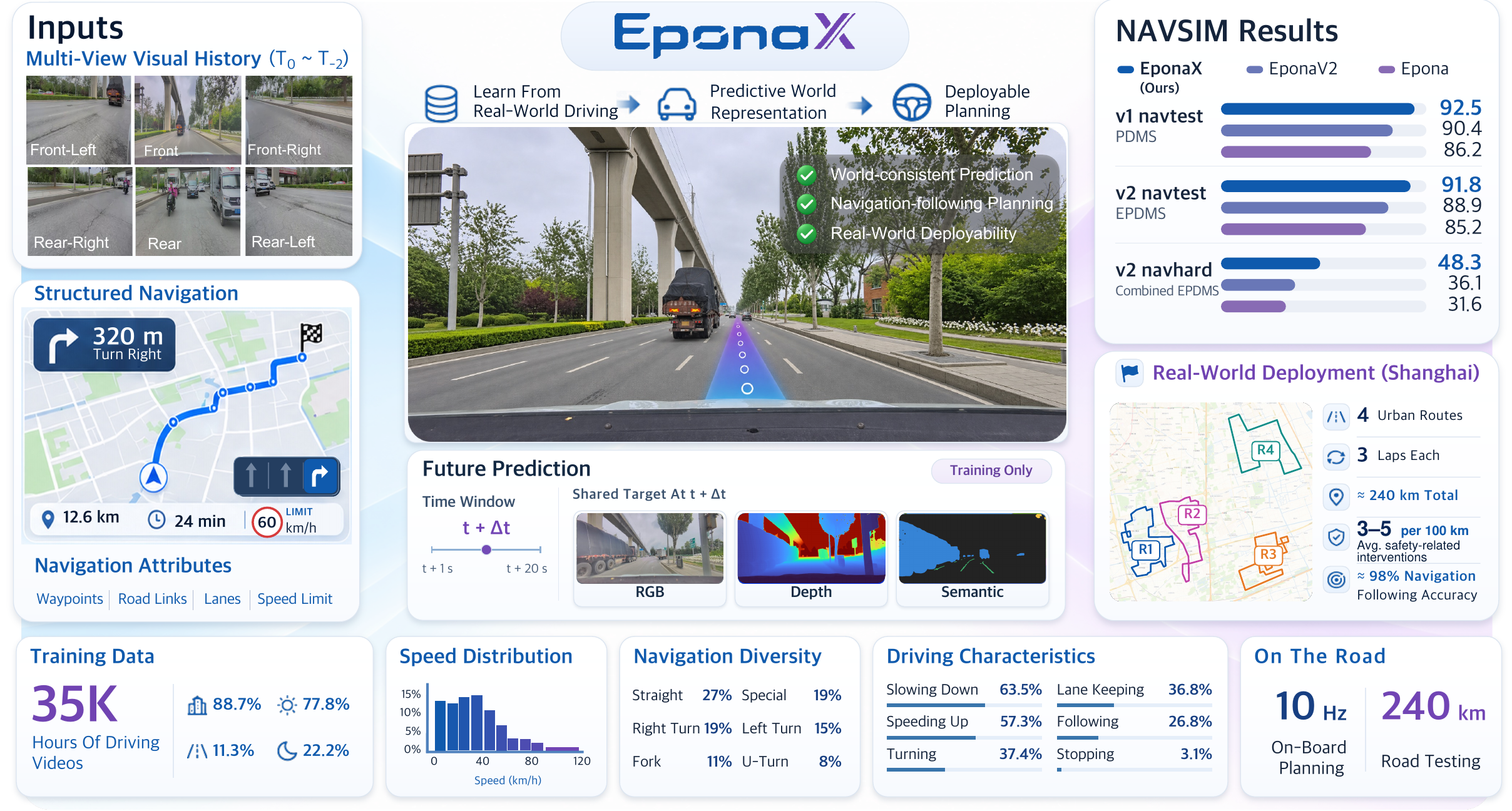}
    \captionof{figure}{\textbf{\model~at a glance.} On the left, multi-view camera history and structured navigation provide scene and route context. In the center, future RGB, depth, and semantics supervise training on 35K hours of driving video, with prediction horizons up to 20\,s. On the right, we report NAVSIM results and summarize road testing in Shanghai, where onboard planning runs at 10\,Hz over approximately 240\,km across three laps on each of four urban routes.}
    \label{fig:eponax_teaser}
\end{minipage}\par
\vspace{14pt}

\reportabstract
\restoregeometry

\setcounter{tocdepth}{2}
\begingroup
\hypersetup{linkcolor=reportAccent,linktoc=all}
\tableofcontents
\endgroup
\clearpage

\setlength{\emergencystretch}{1em}
\section{Introduction}
\label{sec:intro}

Autonomous driving requires a model to interpret current observations and anticipate how the scene may evolve under candidate ego motions. Recent world-action models connect these capabilities by learning predictive scene representations for trajectory planning~\cite{wang2024drivewm,zhang2025epona,xu2026eponav2,li2025drivevlaw0,bartoccioni2025vavim_vavam,zheng2024genad}. Scaling these models from offline benchmarks to real vehicles requires integrating observations from multiple cameras, route constraints, and evidence about future scene dynamics. Both the representation and the trajectory planner must fit within the computation budget for real-time trajectory generation on the vehicle.

Many generative driving models use latent spaces designed for image reconstruction~\cite{gao2024magicdrive-v2,wang2024drivewm,wang2024drivedreamer,hu2024drivingworld,zhang2025epona,russell2025gaia}. Features from vision foundation models can also serve as visual tokens~\cite{zheng2026rae}. DINO-Tok combines deep semantic features with shallow appearance information to support reconstruction~\cite{jia2026dinotok}. EponaV2 uses this representation to jointly learn future appearance, geometry, semantics, and trajectory planning~\cite{xu2026eponav2}. Extending dense visual features~\cite{oquab2024dinov2,simeoni2025dinov3} across multiple cameras and time steps increases the token count. Compact action representations and diffusion planners reduce the cost of action generation~\cite{liao2025diffusiondrive,jia2026walt,pertsch2025fast,liang2026vital}. A deployable world-action model must compactly encode evidence across cameras and time and give the planner access to useful features across the backbone hierarchy as its trajectory estimate evolves.

\begin{figure}[t]
    \centering
    \includegraphics[width=\textwidth]{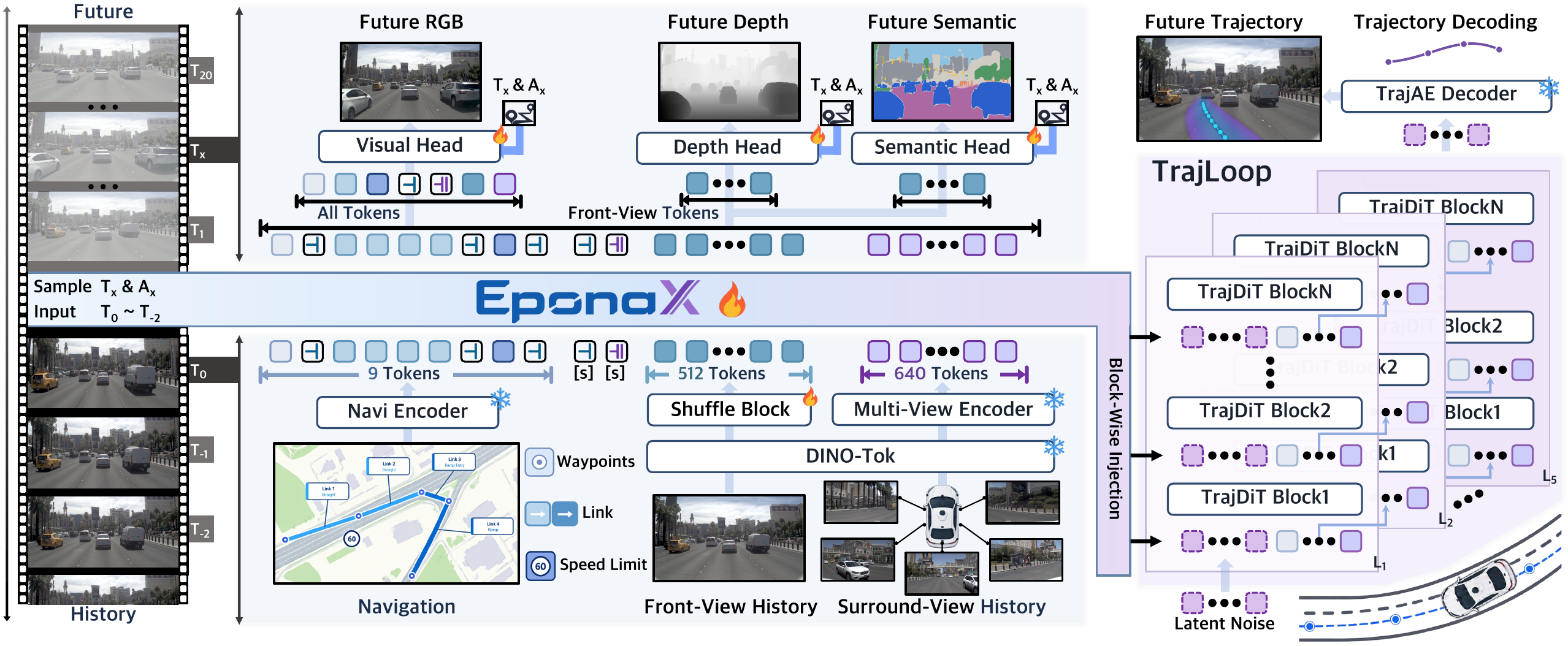}
    \caption{\textbf{Architecture of \model.}
    At the bottom, frozen encoders tokenize camera history and structured navigation for the causal backbone.
    At the top, the RGB, depth, and semantic heads predict observations up to 20\,s ahead during training.
    On the right, TrajLoop reads the corresponding backbone states and updates each planning block's condition at every flow step. A frozen TrajAE decoder maps the resulting latents to a trajectory in the ego frame.}
    \label{fig:framework_overview}
\end{figure}

We present \model, a world-action model designed for onboard planning from visual history and route guidance. As shown in Fig.~\ref{fig:framework_overview}, its asymmetric visual tokenizer preserves detailed front-view features and compresses the five calibrated surround views by a factor of 16. A navigation encoder converts route geometry, maneuvers, lane guidance, and speed limits into semantically aligned tokens. A shared temporal backbone integrates the visual and navigation sequences. Predictions of appearance, depth, and semantics at sampled horizons of up to 20 seconds supervise the backbone during training. On the vehicle, the trajectory pathway uses this shared representation to generate plans at each control update.

\model~retains the latest frame's hidden sequence from each backbone layer and pairs it with the corresponding trajectory interaction block. A frozen WALT trajectory autoencoder~\cite{jia2026walt} encodes waypoints as compact latent tokens for iterative generation. \textbf{TrajLoop} updates each block's scene condition across flow steps, so the condition at each step reflects how the evolving trajectory hypothesis has interacted with that layer's features. Trained on approximately 35K hours of driving video, \model~achieves 92.5 PDMS on NAVSIM v1 and 91.8 EPDMS on NAVSIM v2 navtest after GRPO. On NAVSIM v2 navhard, it achieves a combined 48.3 EPDMS. The policy runs onboard at 10\,Hz over approximately 240\,km across four urban routes in Shanghai, averaging 3--5 safety-related driver interventions per 100\,km.

\noindent\textbf{Contributions}
\begin{itemize}
    \item We introduce \model, a world-action model trained on approximately \textbf{35K hours} of driving data. It uses frozen features from a vision foundation model to encode six camera views and runs its trajectory planner on a real vehicle at \textbf{10\,Hz} with structured route guidance.

    \item We develop a representation that combines asymmetric tokenization of front and surround cameras with semantically aligned navigation. Future appearance, depth, and semantic prediction supervise the shared temporal backbone at sampled horizons of up to 20 seconds.

    \item We develop \textbf{TrajLoop} to generate trajectories in a compact latent space from shared scene features. Each planning block reads the latest frame from its corresponding backbone layer and uses the current trajectory estimate to refine its local condition over successive flow steps.

    \item We evaluate \model~on three NAVSIM splits and over \textbf{240\,km} of road testing across four routes in Shanghai. The road tests cover daytime and nighttime driving, rain, and highway segments.
\end{itemize}

\section{Related Work}
\label{sec:related_work}

\noindent\textbf{World Models for Autonomous Driving.} Driving world models predict scene evolution from historical observations to support simulation and planning~\cite{review4,review5,review6,review7}. Recent work extends the duration of driving video generation~\cite{gao2024magicdrive-v2,gao2024vista,infinitydrive,hu2023gaia-1,jia2023adriver-i}, models scene geometry in 3D and 4D~\cite{uniscene,drivingsphere,liang2026lidarcrafter,xu2025U4D,bian2025dynamiccity,zhu2025spiral}, and synthesizes safety-critical scenarios~\cite{cosmos,terasim,sun2025terasim}. WorldLens evaluates visual fidelity, geometry, and control~\cite{liang2026worldlens}. Feng and Ao measure scene response to a plan separately from camera control~\cite{feng2026camerascene}. Vista uses motion-aware objectives~\cite{gao2024vista}, DriVerse combines trajectory prompting with motion alignment~\cite{driverse}, and ReSim incorporates non-expert behavior collected in a driving simulator~\cite{yang2025resim}.

DrivingGPT and Epona encode actions for controllable autoregressive rollouts~\cite{chen2024drivinggpt,zhang2025epona}. Latent world models also learn predictive features for downstream planning~\cite{li2024law,zheng2025world4drive,yang2025worldrft,driveworld,shu2026,zheng2024occworld,xu2026planwam}. \model~uses reconstructable features and assigns more tokens per camera to the front view, preserving its spatial detail while retaining a compact visual history from all five calibrated surround cameras.

\noindent\textbf{End-to-End Autonomous Driving.} End-to-end driving maps sensor observations to trajectories~\cite{review1,review2,hagedorn2024integration,hu2025vision}. UniAD cascades perception, prediction, and planning~\cite{hu2023uniad}, and PARA-Drive organizes these tasks in parallel~\cite{paradrive}. TransFuser and FusionAD use learned sensor fusion for planning~\cite{chitta2022transfuser,fusionad}. Many planners use dense bird's-eye view (BEV) features~\cite{li2024bevformer,li2024ego,liao2025diffusiondrive,gui2025trajdiff,zhao2026united}. Vectorized scenes and sparse queries offer more compact representations~\cite{jiang2023vad,sparsead,sparsedrive,jia2025drivetransformer,li2025ssr}. VADv2 introduces a trajectory vocabulary~\cite{chen2024vadv2}, Hydra-MDP distills rule-based planning knowledge~\cite{li2024hydra}, and DiffusionDrive truncates trajectory diffusion~\cite{liao2025diffusiondrive}. WoTE evaluates candidate trajectories using future states from a BEV world model~\cite{li2025wote}.

Vision-language-action models connect image and language inputs to robot actions~\cite{kimopenvla,zitkovich2023rt,black2024pi_0,black2025pi_05,wang2025unified}. Driving systems use language models to predict actions~\cite{mao2023gpt,shao2024lmdrive,xu2024drivegpt4,tang2026pixelpilot,wang2025alpamayo,lu2026onevl} and vision-language models for scene understanding and planning~\cite{zhou2024embodied,huang2024drivemm,zhou2025opendrivevla}. DiffVLA couples language guidance with diffusion planning~\cite{jiang2025diffvla}. ORION uses generative action prediction~\cite{fu2025orion}, ReCogDrive adds hierarchical reasoning~\cite{li2025recogdrive}, and DriveVLA-W0 learns features for its action expert through world model pretraining~\cite{li2025drivevlaw0}. \model~pairs each trajectory generation block with its corresponding backbone layer and updates the conditioning tokens in each block as the trajectory estimate evolves across successive flow steps.

\noindent\textbf{Joint World Modeling and Planning.} Intermediate reasoning connects observations to actions in language and multimodal models~\cite{wei2022chain,chen2025towards,wang2025multimodal,xia2025beyond}. ECoT grounds textual reasoning in physical observations~\cite{zawalski2024ecot}, and OneTwoVLA adapts reasoning to control demands~\cite{lin2025onetwovla}. Visual chain-of-thought methods condition actions on generated visual states~\cite{zhao2025cot,lv2025f1}. Driving systems use multimodal reasoning for trajectory prediction~\cite{hwang2024emma,wang2025omnidrive,xing2025openemma,zhou2025autovla}, and AutoDrive-$R^2$ trains the planner to reflect on its predictions~\cite{yuan2025autodrive}.

World models also provide predictions for learning or evaluating plans~\cite{xie2026safedrivevla,li2024think2drive,yang2025raw2drive,yan2026adr1}. Drive-WM evaluates plans using action-conditioned futures~\cite{wang2024drivewm}, and FSDrive uses future images for intermediate reasoning~\cite{zeng2025fsdrive}. DrivingGPT, Epona, EponaV2, PWM, DriveVLA-W0, and DriveLaW jointly learn future prediction and planning~\cite{chen2024drivinggpt,zhang2025epona,xu2026eponav2,zhao2025pwm,li2025drivevlaw0,xia2026drivelaw,zheng2024doe,li2025imagidrive}. SV-WAM uses future video to supervise surround-view planning during training~\cite{wang2026svwam}. Building on EponaV2, \model~uses reconstructable DINO features and trains the shared backbone with appearance, depth, and semantic targets. The deployed planner reads that representation directly to generate trajectories from visual history and route guidance.

\section{\model~Architecture}
\label{sec:bench}

As shown in Fig.~\ref{fig:framework_overview}, \model~uses a shared temporal backbone to integrate multi-view camera history and structured navigation for trajectory planning. Starting from the planning task in Sec.~\ref{sec:method_overview}, we describe how the visual and navigation inputs are tokenized and integrated over time in Secs.~\ref{sec:method_tokenization}--\ref{sec:world_state_modeling}. The resulting representation conditions trajectory generation with TrajLoop in Sec.~\ref{sec:method_trajae_diffusion}, and the scene prediction objectives in Sec.~\ref{sec:method_future_prediction} provide auxiliary supervision for the shared backbone during training.

\subsection{Planning Setup}
\label{sec:method_overview}

At planning time $t$, the model receives $T$ observations at times $t_1<\cdots<t_T=t$. Each observation comprises a front-view image $\mathbf{I}_{t_i}^{\mathrm{f}}$, images $\{\mathbf{I}_{t_i}^{v}\}_{v=1}^{V}$ from $V$ surround cameras, and aligned navigation guidance $\mathbf{n}_{t_i}$. The encoders and backbone, denoted by $F_\theta$, use these observations and the camera calibration $\mathcal{P}_{\mathrm{cam}}$ to compute the feature hierarchy $\mathcal{H}_t$ for the latest frame. The planner $G_\phi$ then predicts a trajectory $\widehat{\mathbf{Y}}_t$ of $P$ poses, each with a planar position and heading in the ego frame.
\begin{equation}
\begin{aligned}
    \mathcal{H}_t
    &=
    F_{\theta}\!\left(
        \left\{
        \mathbf{I}_{t_i}^{\mathrm{f}},
        \{\mathbf{I}_{t_i}^{v}\}_{v=1}^{V},
        \mathbf{n}_{t_i}
        \right\}_{i=1}^{T},
        \mathcal{P}_{\mathrm{cam}}
    \right),\\
    \widehat{\mathbf{Y}}_t
    &=
    G_{\phi}(\mathcal{H}_t)
    \in\mathbb{R}^{P\times3}.
\end{aligned}
\end{equation}

On the vehicle, we use $T=3$ observations and $V=5$ surround cameras. During training, the recorded future trajectory provides the planning target. The future prediction heads use the recorded ego displacement over the sampled prediction horizon as a conditioning input, as described in Sec.~\ref{sec:method_future_prediction}.

\begingroup
\setlength{\intextsep}{12pt plus 1pt minus 1pt}
\begin{figure}[!ht]
    \centering
    \includegraphics[width=\linewidth]{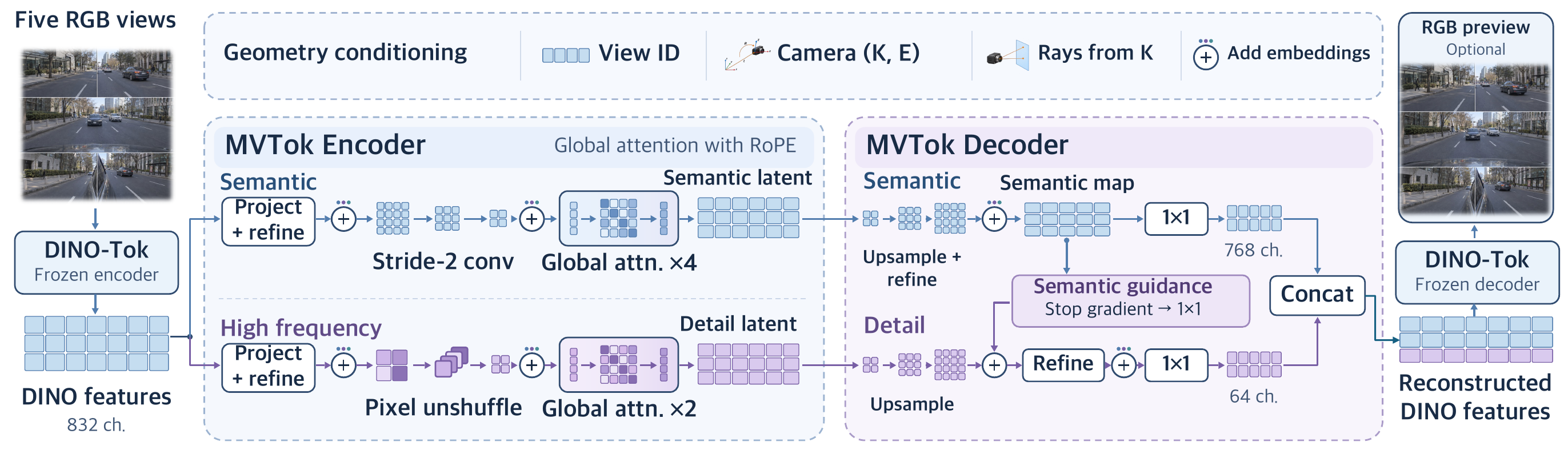}
    \caption{\textbf{Geometry-aware multi-view tokenization.} The encoder compresses semantic and detail DINO-Tok features from five synchronized surround views. During pretraining, the semantic decoder provides a detached feature map to guide detail reconstruction. The frozen DINO-Tok decoder can render RGB previews from the reconstructed features in camera order.}
    \label{fig:fig:mvtok_details}
\end{figure}
\endgroup

\subsection{Multi-View Tokenization}
\label{sec:method_tokenization}

\noindent\textbf{Front-View Encoding.}
To cover more views within a fixed token budget, we preserve front-view detail and compress calibrated surround views. The frozen DINO-Tok encoder $\mathcal{E}_{\mathrm{DINO}}$~\cite{jia2026dinotok} produces $\mathbf{Z}_{t_i}^{\mathrm{f}}=\mathcal{E}_{\mathrm{DINO}}(\mathbf{I}_{t_i}^{\mathrm{f}})\in\mathbb{R}^{N_{\mathrm{f}}\times C_{\mathrm{f}}}$, where $N_{\mathrm{f}}=H_{\mathrm{f}}W_{\mathrm{f}}$. The features combine deep semantics and shallow appearance detail and serve as historical inputs and targets for future visual prediction. Section~\ref{sec:world_state_modeling} describes the token shuffle that shortens the front-view sequence before it enters the temporal backbone.

\noindent\textbf{Surround-View Compression.}
As shown in Fig.~\ref{fig:fig:mvtok_details}, MVTok projects and refines DINO-Tok's semantic and shallow appearance features in separate branches. For view $v$ at time $t_i$, let $\mathbf{S}^{v}$ and $\mathbf{D}^{v}$ denote the resulting grids. The two branches reduce these grids to a common spatial resolution $H_{\mathrm{c}}\times W_{\mathrm{c}}$.
\begin{equation}
    \mathbf{U}_{\mathrm{s}}^{v} = \operatorname{StridedConv}(\mathbf{S}^{v}),
    \qquad
    \mathbf{U}_{\mathrm{d}}^{v} = \operatorname{PixelUnshuffle}(\mathbf{D}^{v}).
    \label{eq:mvtok-compression}
\end{equation}
Strided convolution aggregates semantic context. Pixel unshuffle packs neighboring spatial samples into separate channels according to their offsets within each local block of the input grid.

\Needspace{5\baselineskip}
\noindent\textbf{Geometry-Aware Fusion.}
For each branch $b\in\{\mathrm{s},\mathrm{d}\}$, we stack the reduced grids $\mathbf{U}_{b}=(\mathbf{U}_{b}^{v})_{v=1}^{V}$ in camera order. We add view, calibration, and ray embeddings at successive resolutions. The cross-view attention operator $\mathcal{A}_{b}$ uses rotary encodings of view and spatial indices. Rays specify directions in the camera frame, and the calibration embedding encodes intrinsics and ego-relative extrinsics. We refine each branch independently and concatenate their outputs at each spatial position along the channel axis, denoted by $\Vert$.
\begin{equation}
    \mathbf{Z}_{t_i}^{\mathrm{mv}}
    =
    \bigl[
      \mathcal{A}_{\mathrm{s}}(\mathbf{U}_{\mathrm{s}},\mathcal{P}_{\mathrm{cam}})
      \,\Vert\,
      \mathcal{A}_{\mathrm{d}}(\mathbf{U}_{\mathrm{d}},\mathcal{P}_{\mathrm{cam}})
    \bigr]
    \in\mathbb{R}^{N_{\mathrm{mv}}\times C_{\mathrm{mv}}}.
    \label{eq:mvtok-fusion}
\end{equation}
Here $N_{\mathrm{mv}}=VH_{\mathrm{c}}W_{\mathrm{c}}$ and $C_{\mathrm{mv}}$ is the combined channel width. With five surround cameras, MVTok yields 640 spatial tokens, a $16\times$ reduction. Section~\ref{sec:mvtok_pretraining} describes the separate reconstruction objectives used to pretrain the tokenizer as a deterministic frame-wise autoencoder. We then freeze the encoder for historical inputs and future targets, using the same camera order and spatial layout for both sequences.

\begingroup
\setlength{\intextsep}{12pt plus 1pt minus 1pt}
\begin{figure}[!htp]
    \centering
    \includegraphics[width=\linewidth]{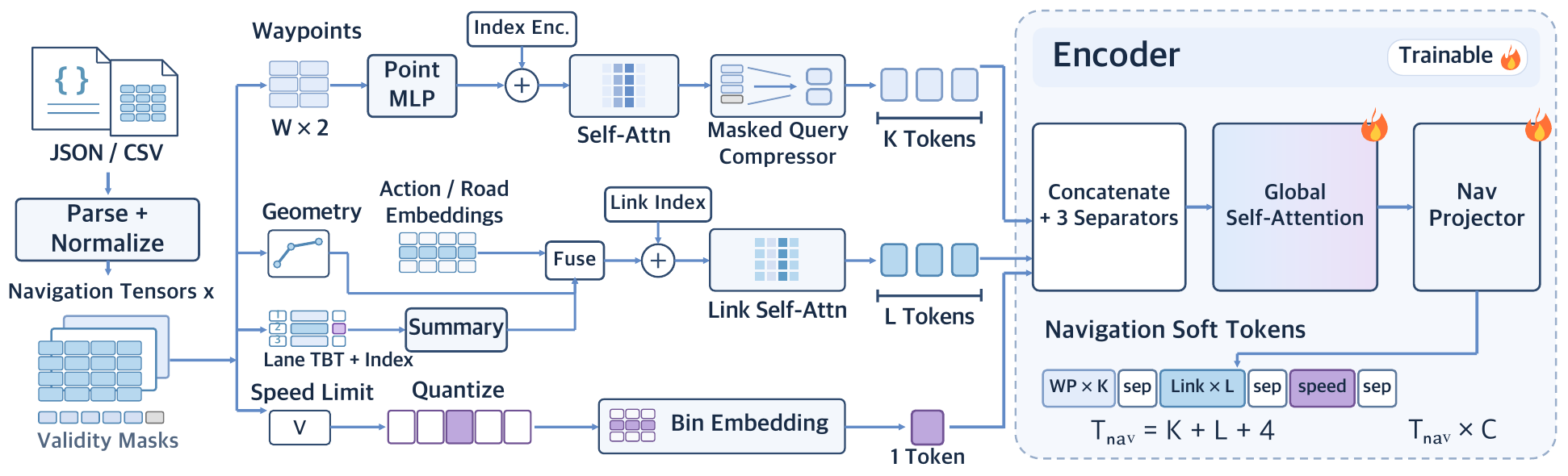}
    \caption{\textbf{Structured navigation encoding.}
    The three branches encode waypoints, links with lane attributes, and speed limits for global self-attention and residual projection. The diagram labels the waypoint count $K$ and the link slot count $L$. The sequence has $K+L+4$ positions, including one speed token and three separators. Navigation alignment trains the encoder and projector. Joint pretraining and deployment use the resulting frozen modules.}
    \label{fig:navigation_structure}
\end{figure}
\endgroup

\subsection{Structured Navigation Encoding}
\label{sec:method_navigation}

Route intent comes from waypoints in the ego frame, ordered road links with maneuver and lane attributes, and speed limits. As shown in Fig.~\ref{fig:navigation_structure}, the encoder represents these fields in three branches. Validity masks identify the waypoints, road links, and lanes available in each route record.

\noindent\textbf{Waypoint branch.}
A pointwise projection embeds the 2D waypoints in the ego frame. Index embeddings preserve their order, and self-attention relates points along the route. A learned query pools valid features into one waypoint token, so several points along a bend can inform that token.

\noindent\textbf{Road-link and lane branch.}
For each upcoming road link, we embed its start position, relative displacement, distance, and direction, together with its primary maneuver, auxiliary maneuver, and road type. The first link uses the ego vehicle as its reference, and subsequent links use their predecessors. Within each link, we use lane indices to preserve lane order and aggregate lane actions and recommendations. We fuse the lane summary with the link embedding. Self-attention across the ordered link tokens then captures sequences of maneuvers, such as a lane change followed by a turn at an upcoming intersection.

\noindent\textbf{Speed and fusion.}
At time $t_i$, the branch outputs are the waypoint token $\mathbf{w}$, the ordered link slots $(\boldsymbol{\ell}_{j})_{j=1}^{J}$, and a token $\mathbf{u}$ for the discretized speed limit. A learned separator $\mathbf{q}_b$ follows each branch. Global self-attention $\mathcal{A}_{\mathrm{nav}}$ fuses the sequence, and the residual projector $\mathcal{P}_{\mathrm{nav}}$ maps it to width $C$.
\begin{equation}
\begin{aligned}
    \mathbf{Z}_{t_i}^{\mathrm{nav}}
    &= [\mathbf{w},\mathbf{q}_{\mathrm{wp}},
             \boldsymbol{\ell}_1,\ldots,\boldsymbol{\ell}_J,
             \mathbf{q}_{\mathrm{link}},\mathbf{u},\mathbf{q}_{\mathrm{spd}}],\\
    \overline{\mathbf{Z}}_{t_i}^{\mathrm{nav}}
    &= \mathcal{P}_{\mathrm{nav}}\!\left(
        \mathcal{A}_{\mathrm{nav}}(\mathbf{Z}_{t_i}^{\mathrm{nav}})
      \right)
    \in \mathbb{R}^{N_{\mathrm{nav}}\times C}.
\end{aligned}
\label{eq:navigation-encoding}
\end{equation}
The sequence has $N_{\mathrm{nav}}=J+5$ positions and carries a padding mask. With $J=4$ link slots, we obtain nine navigation tokens per frame. Section~\ref{sec:world_state_modeling} describes how we add boundary tokens before the backbone.

We pretrain the encoder and projector to predict navigation text with a frozen language model and reconstruct link geometry, maneuver and lane attributes, and speed limits. These objectives align the tokens with navigation descriptions and preserve the underlying route fields. We freeze both modules after alignment. Section~\ref{sec:navigation_alignment} gives the input specifications, objective, and reconstruction check.

\subsection{Shared Temporal World Representation}
\label{sec:world_state_modeling}

A learnable token shuffle shortens the front-view sequence to $N_{\mathrm{f}}'$ tokens. We project the front and surround features to width $C$, giving $\overline{\mathbf{Z}}_{t_i}^{\mathrm{f}}$ and $\overline{\mathbf{Z}}_{t_i}^{\mathrm{mv}}$. The frame sequence $\mathbf{X}_{t_i}$ places the navigation sequence first, followed by learned tokens $\mathbf{s}^{\mathrm{nav}}$ and $\mathbf{s}^{\mathrm{img}}$ that mark the route boundary and the start of the visual sequence.
\begin{equation}
    \mathbf{X}_{t_i}
    =
    [\overline{\mathbf{Z}}_{t_i}^{\mathrm{nav}},
     \mathbf{s}^{\mathrm{nav}},
     \mathbf{s}^{\mathrm{img}},
     \overline{\mathbf{Z}}_{t_i}^{\mathrm{f}},
     \overline{\mathbf{Z}}_{t_i}^{\mathrm{mv}}]
    \in\mathbb{R}^{N\times C},
    \qquad N=N_{\mathrm{nav}}+N_{\mathrm{f}}'+N_{\mathrm{mv}}+2.
    \label{eq:temporal-frame-tokens}
\end{equation}

The Qwen3-VL 2B language transformer~\cite{bai2025qwen3vl} applies causal attention with its native positional encoding to the chronological sequence $\mathbf{X}=[\mathbf{X}_{t_1},\ldots,\mathbf{X}_{t_T}]$. At layer $\ell$, $\mathbf{H}^{(\ell)}\in\mathbb{R}^{TN\times C}$ contains the hidden states for all $T$ frames. The hierarchy $\mathcal{H}_t$ retains the rows for the latest frame from all $L$ layers.
\begin{equation}
    \mathcal{H}_t
    =
    (\mathbf{H}_t^{(\ell)})_{\ell=1}^{L},
    \qquad
    \mathbf{H}_t^{(\ell)}
    = \mathbf{H}^{(\ell)}_{(T-1)N+1:TN,:}
    \in\mathbb{R}^{N\times C}.
    \label{eq:temporal-layer-readout}
\end{equation}

We use $L=28$ backbone layers. A $2\times2$ shuffle reduces $N_{\mathrm{f}}=2048$ front-view tokens to $N_{\mathrm{f}}'=512$. The $N_{\mathrm{mv}}=640$ surround tokens join the compressed front-view sequence. Adding $N_{\mathrm{nav}}=9$ navigation tokens and two learned boundary tokens yields $N=1163$ positions in each frame sequence.

\subsection{Trajectory Planning with Recurrent Conditions}
\label{sec:method_trajae_diffusion}
\label{app:trajloop_unroll}

A frozen WALT trajectory autoencoder~\cite{jia2026walt} encodes demonstrated motion in compact latents aligned with the world model. These latents contain reconstruction and semantic channels. TrajDiT~\cite{peebles2023scalable} generates them by conditional flow matching and pairs each of its $L$ blocks with a backbone layer.

The backbone runs once on the observations. A shared linear projection $\mathcal{P}_{\mathrm{c}}$ initializes the condition of block $i$ as $\mathbf{c}_i^{(0)}=\mathcal{P}_{\mathrm{c}}(\mathbf{H}_t^{(i)})$. The fixed-condition baseline uses $\mathbf{c}_i^{(r)}=\mathbf{c}_i^{(0)}$ at every sampling step.

At flow time $\tau_r$, TrajDiT takes a trajectory latent $\mathbf{x}^{(r)}$ and conditions $\mathcal{C}^{(r)}=(\mathbf{c}_i^{(r)})_{i=1}^{L}$ to predict the flow velocity $\mathbf{v}^{(r)}$. Within block $\mathcal{B}_i$, joint attention updates the incoming trajectory tokens $\mathbf{u}_i^{(r)}$ and the local condition. Trajectory tokens pass to the next block within the step. The full condition output, including residual updates, stays with block $i$ at the same token count for the next flow step.
\begin{equation}
\begin{aligned}
\bigl(\mathbf{u}_{i+1}^{(r)},\mathbf{c}_i^{(r+1)}\bigr)
&=\mathcal{B}_i\bigl(\mathbf{u}_i^{(r)},\mathbf{c}_i^{(r)},\tau_r\bigr),\\
(\mathbf{v}^{(r)},\mathcal{C}^{(r+1)})
&=\operatorname{TrajDiT}_{\phi}
(\mathbf{x}^{(r)},\mathcal{C}^{(r)},\tau_r),\\
\mathbf{x}^{(r+1)}
&=\mathbf{x}^{(r)}
+(\tau_{r+1}-\tau_r)\mathbf{v}^{(r)},
\end{aligned}
\label{eq:trajloop_update}
\end{equation}

Both planners start from Gaussian noise and use $R=5$ Euler steps at inference. TrajLoop unrolls the same $R$ steps during training, with gradients through both the trajectory and condition states. Section~\ref{sec:joint_pretraining} gives the objective that supervises each flow prediction and the terminal latent.

The frozen decoder maps the final $N_{\mathrm{traj}}$ latent tokens in $\mathbf{x}^{(R)}$ to the $P$ poses of $\widehat{\mathbf{Y}}_t$. The onboard configuration uses $N_{\mathrm{traj}}=10$ and $P=60$ for a six-second trajectory in the ego frame.

\subsection{Future Scene Predictions}
\label{sec:method_future_prediction}

For a sampled horizon $\Delta t$, the future targets come from the recorded observations at $t+\Delta t$. We compute the relative ego displacement $\mathbf{a}_{t,\Delta t}=(\Delta x,\Delta y,\Delta\theta)$ from the recorded poses to account for the change in viewpoint. Each future head $j$ draws on a selection $\mathcal{R}_j$ of the shared features and receives a time embedding $e_j(\Delta t)$ and the displacement after the backbone processes the observation history.
\begin{equation}
    \mathbf{c}_{j,t,\Delta t}
    = \bigl(\mathcal{R}_j(\mathcal{H}_t),\,e_j(\Delta t),\,\mathbf{a}_{t,\Delta t}\bigr),
    \qquad j\in\{\mathrm{vis},\mathrm{depth},\mathrm{sem}\}.
    \label{eq:future-head-condition}
\end{equation}
Future visual prediction uses the full final sequence $\mathbf{H}_t^{(L)}$, and the depth and semantic heads select its front-view positions. For current depth and semantics, the heads read front-view positions from task-specific intermediate layers. We sample $\Delta t$ from $\{1,\ldots,20\}\,\mathrm{s}$. Section~\ref{sec:joint_pretraining} details target construction and the loss terms used to supervise the shared backbone through each prediction head.

\noindent\textbf{Future Visual Prediction.}
Following Epona~\cite{zhang2025epona}, conditional latent flow matching predicts synchronized front-view and surround-view features encoded by the frozen tokenizers. The objective $\mathcal{L}_{\mathrm{vis}}$ balances semantic and high-frequency channels across both view groups. The target sequence keeps a fixed order across cameras and matches each token to its spatial position within the corresponding camera grid.

\noindent\textbf{Temporal Depth Prediction.}
A frozen Depth Anything 3 teacher~\cite{lin2025depth} provides pseudo-labels for a shared head that predicts current and future front-view depth. Current targets supervise observed geometry, and future targets supervise geometry at the sampled time and viewpoint. The objective $\mathcal{L}_{\mathrm{depth}}$ combines confidence-weighted log-depth reconstruction with spatial gradient matching for both predictions.

\noindent\textbf{Temporal Semantic Prediction.}
A shared head takes semantic prompts and learns current and future front-view semantics from frozen SAM3 targets~\cite{carion2026sam}. The objective $\mathcal{L}_{\mathrm{sem}}$ aligns both predictions with the teacher's targets, balances positive and negative regions, and provides contrastive supervision.

\noindent\textbf{Inference.}
The deployed system runs the visual and navigation encoders, shared backbone, TrajLoop, and frozen trajectory decoder to produce trajectories in the ego frame at each planning step.

\section{Training Recipe}
\label{sec:training_recipe}

We pretrain the surround-view tokenizer and align the navigation encoder with route descriptions. With both encoders frozen, we jointly train the backbone, planner, and scene prediction heads on driving video, then adapt the planner to NAVSIM through supervised fine-tuning and Flow-GRPO.

\subsection{Multi-View Tokenizer Pretraining}
\label{sec:mvtok_pretraining}

\noindent\textbf{Feature Reconstruction.}
MVTok reconstructs frozen DINO-Tok features from synchronized surround views. As shown in Fig.~\ref{fig:fig:mvtok_details}, the semantic decoder reconstructs scene features, and a projection of its detached output guides detail reconstruction. The feature MSEs $\mathcal{L}_{\mathrm{sem}}^{\mathrm{rec}}$ and $\mathcal{L}_{\mathrm{hf}}^{\mathrm{rec}}$ supervise the semantic and detail branches. Spatial gradient matching supervises local boundaries, a pooled and detached DINO target guides the semantic latent, and a channelwise moment penalty regularizes both latent groups.
\begin{equation}
    \mathcal{L}_{\mathrm{MVTok}}
    = \mathcal{L}_{\mathrm{sem}}^{\mathrm{rec}}
    + 4\,\mathcal{L}_{\mathrm{hf}}^{\mathrm{rec}}
    + 0.5\,\mathcal{L}_{\mathrm{grad}}
    + 0.1\,\mathcal{L}_{\mathrm{distill}}
    + 0.001\,\mathcal{L}_{\mathrm{reg}}.
    \label{eq:mvtok-pretraining-objective}
\end{equation}

\noindent\textbf{Training Setup.}
We jointly train the encoder, decoder, geometry embeddings, and auxiliary semantic head with DINO-Tok frozen. The undistortion-aware stage resumes from an earlier MVTok model with a fresh optimizer. Tab.~\ref{tab:multiview_training_config} gives the feature grids, parameter counts, and selected checkpoint. The temporal backbone uses the frozen encoder for historical inputs and future targets.

\begin{trainingconfiguration}
    {\textbf{MVTok pretraining setup.} Eq.~\eqref{eq:mvtok-pretraining-objective} defines the feature reconstruction objective.}
    {tab:multiview_training_config}
    \begin{configurationgroup}{Inputs and Checkpoint}
        \configurationfield{Input}{5 calibrated surround views per frame}
        \configurationfield{Grid per view}{$32\times64$ to $8\times16$}
        \configurationfield{Spatial tokens}{10,240 to 640 across all views}
        \configurationfield{Checkpoint}{Step 40,000 of the undistortion-aware stage}
    \end{configurationgroup}\hfill
    \begin{configurationgroup}{Model Parameters}
        \configurationfield{Encoder}{67.7M with geometry embeddings}
        \configurationfield{Training model}{104.4M with decoder and auxiliary semantic head}
        \configurationfield{Excluded}{DINO-Tok parameters}
    \end{configurationgroup}
\end{trainingconfiguration}

\noindent\textbf{Reconstruction Check.}
We reconstruct three time steps from one recorded clip, with five calibrated views at each step. Within each feature group, we average squared errors over time, views, spatial positions, and channels in the original DINO-Tok feature space. At a spatial compression ratio of $16\times$, we obtain an MSE of $4.86\times10^{-3}$ for semantic features and $2.21\times10^{-2}$ for the detail features.

\subsection{Navigation Alignment}
\label{sec:navigation_alignment}

\noindent\textbf{Route Inputs and Text Targets.}
We interpolate and clip the waypoint sequence, express waypoints and link start positions in the ego frame, and index discrete map fields. Each frame has up to 64 waypoint samples, four upcoming link slots, and eight ordered lane slots per link. The residual projector maps the 384-dimensional navigation representation to the 2048-dimensional backbone space.

We verbalize the valid route fields and use an offline Qwen3.6~35B teacher~\cite{qwen36_35b_a3b} to rewrite them as route guidance. Prompt variants preserve distances, directions, road types, maneuvers, lane recommendations, and speed limits while varying the wording. We cache each description by frame. Fig.~\ref{fig:navigation_pretraining} shows a straight maneuver in 14.6\,m, a 20\,km/h speed limit, and guidance to use lanes 2--4.

\begingroup
\setlength{\intextsep}{8pt plus 1pt minus 1pt}
\begin{figure}[!htp]
    \centering
    \includegraphics[width=\linewidth]{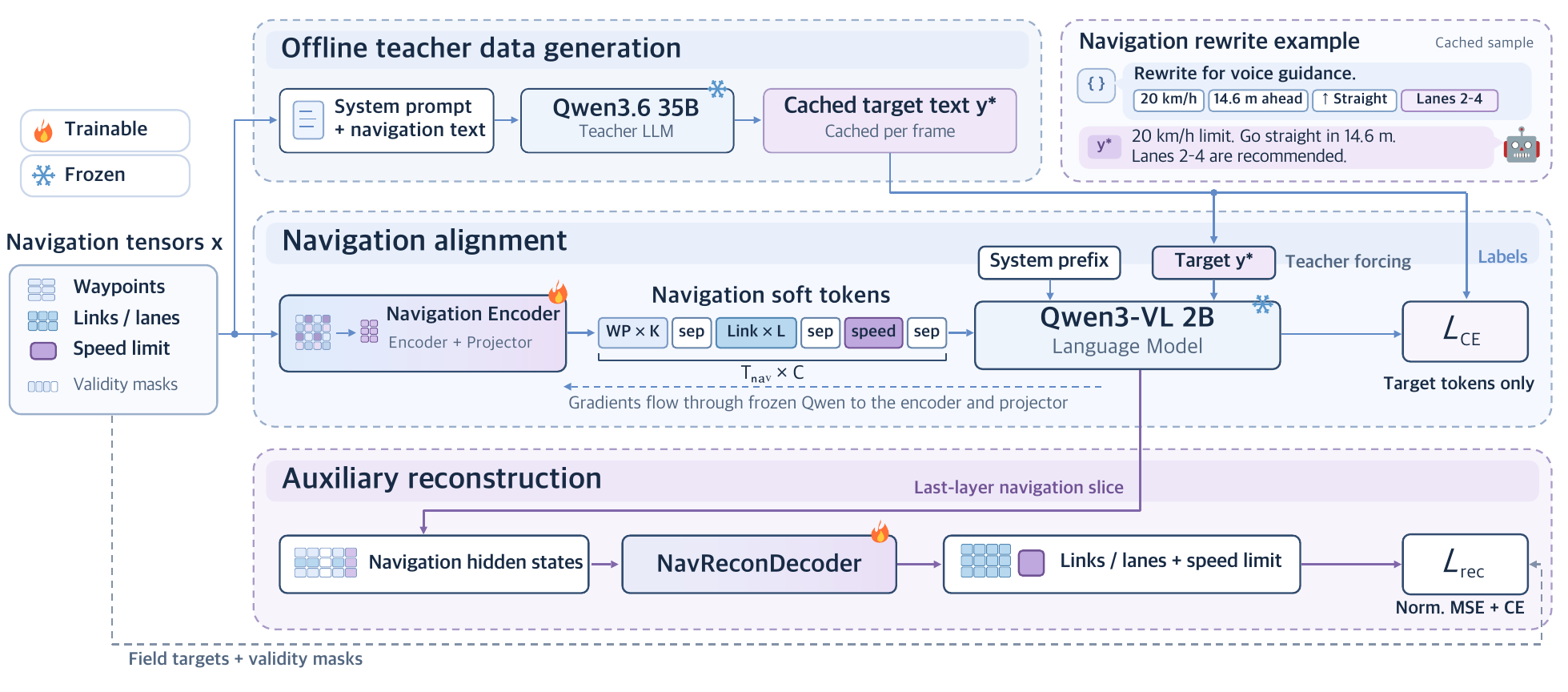}
    \caption{\textbf{Navigation encoder pretraining.} A cached teacher rewrite at the top and field reconstruction at the bottom supervise the projected tokens. Text gradients pass through the frozen alignment model. The diagram uses $T_{\mathrm{nav}}$ for the sequence length $N_{\mathrm{nav}}$ in Sec.~\ref{sec:method_navigation}.}
    \label{fig:navigation_pretraining}
\end{figure}
\endgroup

\noindent\textbf{Alignment Objective.}
A frozen Qwen3-VL~2B language model~\cite{bai2025qwen3vl} receives the system prefix, projected navigation tokens, and target text under teacher forcing. The next-token cross-entropy $\mathcal{L}_{\mathrm{text}}$ applies to target-text positions. Gradients through the language model update the navigation encoder and projector.

A reconstruction decoder reads the final hidden states at the navigation-token positions. It predicts link starts and speed limits with normalized squared error, and maneuver, road type, and lane attributes with cross-entropy. The route waypoints provide geometry to the encoder. We average the seven field reconstruction losses $\ell_j$ with equal weight before adding the text prediction loss.
\begin{equation}
    \mathcal{L}_{\mathrm{nav}}
    = \mathcal{L}_{\mathrm{text}} + \frac{1}{7}\sum_{j=1}^7\ell_j.
    \label{eq:navigation_alignment_loss}
\end{equation}

\noindent\textbf{Training Setup.}
Tab.~\ref{tab:navigation_alignment_config} gives the base alignment configuration. Joint pretraining and deployment use the encoder and projector frozen at the lane guidance checkpoint saved at step 678{,}000.

\begin{trainingconfiguration}
    {\textbf{Navigation alignment recipe.} The hardware and optimizer settings describe the base stage that trains the navigation encoder and projector through the frozen language model.}
    {tab:navigation_alignment_config}
    \begin{configurationgroup}{Data and Supervision}
        \configurationfield{Inputs}{Time-aligned SD navigation and cached teacher rewrites}
        \configurationfield{Sampling}{8 frames per sample, up to 2 descriptions per frame}
        \configurationfield{Tokens}{1 waypoint, 4 link, 1 speed, 3 separators}
        \configurationfield{Objectives}{Target-text cross-entropy and mean field reconstruction loss}
    \end{configurationgroup}\hfill
    \begin{configurationgroup}{Model and Optimization}
        \configurationfield{Parameters}{17.0M for encoder and projector, 20.7M with field decoder}
        \configurationfield{Excluded}{Frozen alignment language model parameters}
        \configurationfield{Hardware}{6 nodes with 8 GPUs each}
        \configurationfield{Effective batch}{48}
        \configurationfield{Optimizer}{Learning rate~$10^{-4}$, weight decay~$10^{-2}$, 500 warm-up steps}
    \end{configurationgroup}
\end{trainingconfiguration}

\noindent\textbf{Reconstruction Check.}
At the selected checkpoint, we evaluate four valid frames from each of three cached clips using the frozen alignment model. We compute the reconstruction errors with training normalization scales of 300 for link coordinates and 140 for speed limits. Across the 12 frames, the normalized MSE is $6.28\times10^{-3}$ for link start coordinates and $1.29\times10^{-2}$ for the speed limit fields.

\subsection{World-Action Pretraining}
\label{sec:joint_pretraining}
\label{app:trajloop_training}
\label{sec:method_training_inference}

\noindent\textbf{Training Data and Modules.}
We jointly train the shared backbone, TrajLoop, and the three scene prediction heads on approximately 35K hours of driving video using 1{,}024 NVIDIA H20 GPUs. The visual and navigation encoders, WALT trajectory autoencoder, and teachers remain frozen. Fig.~\ref{fig:eponax_teaser} shows the speed and navigation distributions. Speeds are mostly below 60~km/h, with a tail beyond 90~km/h.

\noindent\textbf{Joint Objective.}
The trajectory loss trains the planning pathway. The visual, depth, and semantic losses provide auxiliary supervision through the prediction heads introduced in Sec.~\ref{sec:method_future_prediction}. The coefficients $\lambda_j$ set the relative weight of each learning signal during joint world-action pretraining.
\begin{equation}
    \mathcal{L}_{\mathrm{train}}
    =
    \lambda_{\mathrm{traj}}\mathcal{L}_{\mathrm{traj}}
    + \lambda_{\mathrm{vis}}\mathcal{L}_{\mathrm{vis}}
    + \lambda_{\mathrm{depth}}\mathcal{L}_{\mathrm{depth}}
    + \lambda_{\mathrm{sem}}\mathcal{L}_{\mathrm{sem}}.
    \label{eq:world-action-objective}
\end{equation}

\noindent\textbf{Trajectory Supervision.}
The fixed-condition TrajDiT baseline trains at a sampled flow time on an interpolation of Gaussian noise and the reference trajectory latent. TrajLoop starts from one noise sample $\boldsymbol{\epsilon}$ and unrolls the $R=5$ Euler updates defined in Sec.~\ref{sec:method_trajae_diffusion}. We supervise each predicted velocity $\mathbf{v}^{(r)}$ and the final trajectory latent $\mathbf{x}^{(R)}$ using the standardized demonstration latent $\mathbf{z}$.
\begin{equation}
    \mathcal{L}_{\mathrm{traj}}
    =
    d\!\left(\mathbf{x}^{(R)},\mathbf{z}\right)
    + \frac{1}{R}\sum_{r=0}^{R-1}
      d\!\left(\mathbf{v}^{(r)},\mathbf{z}-\boldsymbol{\epsilon}\right).
    \label{eq:trajloop_training}
\end{equation}
The distance $d$ sums the weighted MSEs for the reconstruction and semantic channels. All steps share the same flow target, and the terminal latent receives direct reconstruction supervision. Gradients pass through both recurrent states, so the planner learns the conditions through their effect on the trajectory.

\noindent\textbf{Scene Supervision.}
As defined in Sec.~\ref{sec:method_future_prediction}, the future heads use time and pose conditions after shared feature extraction. Tab.~\ref{tab:scene-supervision} lists each head's parameter count, targets, and training signal.

\begin{trainingconfiguration}
    {\textbf{Scene supervision during joint pretraining.}
    Depth and semantics use front-view targets at $t$ and $t+\Delta t$. The visual head uses synchronized front and surround views at $t+\Delta t$.}
    {tab:scene-supervision}
    \begin{tabularx}{\linewidth}{@{}>{\raggedright\arraybackslash}X@{\hspace{20pt}}>{\raggedright\arraybackslash}X@{}}
        \reporthead{Prediction Heads and Targets} & \reporthead{Training Objectives} \\
        \noalign{\vskip 1pt{\color{reportAccent!24!white}\hrule height 0.4pt}\vskip 2pt}
        \textbf{Visual} (562.0M)\newline Frozen front and surround tokenizers
            & $\mathcal{L}_{\mathrm{vis}}$ Latent flow matching\newline Semantic and appearance channels \\
        \addlinespace[3pt]
        \textbf{Depth} (109.4M)\newline Depth Anything 3~\cite{lin2025depth}
            & $\mathcal{L}_{\mathrm{depth}}$ Confidence-weighted log-depth\newline Reconstruction and spatial gradient matching \\
        \addlinespace[3pt]
        \textbf{Semantics} (58.0M)\newline SAM3~\cite{carion2026sam}
            & $\mathcal{L}_{\mathrm{sem}}$ Prompt-conditioned prediction\newline Balanced positive and negative regions \\
    \end{tabularx}
\end{trainingconfiguration}

For appearance, we concatenate the frozen front-view and MVTok targets in camera and spatial order. The visual head predicts the flow from Gaussian noise to these features. The loss evaluates the semantic and appearance channels separately for both groups of camera views~\cite{zhang2025epona}.

For depth, we rescale teacher targets to a canonical focal length~\cite{yin2023metric3d} before taking their logarithm. Learned confidence downweights uncertain targets. A negative log-confidence penalty discourages uniformly low confidence, and spatial gradient matching supervises geometry at depth boundaries.

For semantics, the teacher provides prompt embeddings, soft targets, and masks for balancing positive and negative regions. Cross-entropy and reverse KL match the teacher's soft targets, and a contrastive term separates regions that match a prompt from difficult negative regions.

\subsection{NAVSIM Adaptation}
\label{sec:navsim_grpo}

\noindent\textbf{Supervised Fine-Tuning.}
We fine-tune the pretrained model on four-frame clips from the NAVSIM \emph{navtrain} split~\cite{bench1,bench2} using its native navigation inputs. The supervised checkpoint initializes Flow-GRPO post-training of the trajectory planner on the same split, following EponaV2~\cite{xu2026eponav2}.

\noindent\textbf{Stochastic Rollouts.}
For each scene, we sample $G$ candidates from independent Gaussian noise with the same initial scene context. Each candidate carries its own recurrent conditions across $R$ stochastic transitions in the standardized WALT latent space~\cite{lipman2022flow,shao2024deepseekmath,liu2026flow}. At flow time $\tau_r$, the Gaussian transition uses a drift-corrected flow update as its mean. The nominal standard deviation is $a\sqrt{(1-\tau_r)\Delta\tau_r/\tau_r}$, where $a$ is the noise coefficient and $\Delta\tau_r$ the step size. The initial transition evaluates the denominator at the first positive flow time. The frozen WALT decoder maps each terminal latent to eight trajectory points over four seconds using the NAVSIM rollout configuration given in Tab.~\ref{tab:flow_grpo_config}.

\noindent\textbf{Reward and Update.}
We score the decoded candidates in a simulated rollout. Let $p$ be terminal progress projected along the demonstrated displacement and normalized by its length, $c$ the rollout comfort score, and $d_{\mathrm{demo}}$ the $L_1$ deviation from the demonstration. The reward $q$ combines these terms.
\begin{equation}
    q = \frac{5p+2c}{7}\exp(-d_{\mathrm{demo}}).
    \label{eq:grpo-reward}
\end{equation}
Within each scene, we center the candidate rewards and divide by their standard deviation plus $10^{-4}$ to form advantages. The policy loss $\mathcal{L}_{\mathrm{RL}}$ averages negative transition log densities over candidates and steps with advantage weights and discount $\gamma^{R-1-r}$. We clip elementwise log densities before averaging over latent tokens and channels. Advantages and sampled next states are fixed during differentiation.

The imitation loss $\mathcal{L}_{\mathrm{IL}}$ uses the same rollout states. Its target velocity is the difference between the standardized demonstration and detached current latents, divided by the remaining flow time floored at $10^{-5}$. We sum mean absolute velocity errors for the reconstruction and semantic channels.
\begin{equation}
    \mathcal{L}_{\mathrm{GRPO}}
    = \mathcal{L}_{\mathrm{RL}} + 0.8\,\mathcal{L}_{\mathrm{IL}}.
    \label{eq:grpo_post_training_loss}
\end{equation}

\noindent\textbf{Training Setup.}
TrajDiT is the trainable module during post-training. Tab.~\ref{tab:flow_grpo_config} gives the latent representation, sampling settings, loss weights, and optimizer. We evaluate one trajectory per scene in Sec.~\ref{sec:experiments}.

\begin{trainingconfiguration}
    {\textbf{NAVSIM Flow-GRPO post-training configuration.} Noise scales are defined in standardized latent coordinates. The effective batch counts input scenes before candidate sampling.}
    {tab:flow_grpo_config}
    \begin{configurationgroup}{Trajectory Representation}
        \configurationfield{Trajectory}{8 points over 4~s in 4 latent tokens}
        \configurationfield{Channels}{8 reconstruction and 24 semantic}
        \configurationfield{Scaling floor}{$10^{-6}$ for standardization}
    \end{configurationgroup}\hfill
    \begin{configurationgroup}{Rollout and Scoring}
        \configurationfield{Sampling}{$G=8$ candidates, $R=5$ steps}
        \configurationfield{Noise scale}{$a=0.01$}
        \configurationfield{Std. dev. floors}{0.04 for sampling, 0.1 for scoring}
        \configurationfield{Clipping}{Noise $[-5.0,5.0]$ and elementwise log density $[-5.0,2.0]$}
    \end{configurationgroup}
    \eponaxPanelNextRow
    \begin{configurationgroup}{Post-Training Objective}
        \configurationfield{Loss weights}{1.0 for RL and 0.8 for imitation}
        \configurationfield{Discount}{$\gamma=0.6$}
        \configurationfield{Adv. offset}{$10^{-4}$ in the denominator}
        \configurationfield{Flow target floor}{$10^{-5}$ in the denominator}
    \end{configurationgroup}\hfill
    \begin{configurationgroup}{Optimization}
        \configurationfield{Trainable}{TrajDiT}
        \configurationfield{AdamW}{Learning rate~$5\times10^{-5}$, weight decay~$10^{-2}$}
        \configurationfield{Coefficients}{$(\beta_1,\beta_2)=(0.9,0.999)$ and $\epsilon=10^{-8}$}
        \configurationfield{Gradient clip}{Norm 1.0}
        \configurationfield{Effective batch}{32 scenes on 4 nodes of 8 GPUs}
    \end{configurationgroup}
\end{trainingconfiguration}

\FloatBarrier

\section{Experiments}\label{sec:experiments}

\subsection{Training Data, Evaluation Protocols, and Implementation}

\noindent\textbf{Training data and pretraining.}
We pretrain on nearly 35K hours of real-world driving video using 1,024 NVIDIA H20 GPUs. Fig.~\ref{fig:eponax_teaser} shows the distribution of speeds, navigation commands, and driving behaviors. Most samples are below 60~km/h, with a long tail extending beyond 90~km/h. The recordings include maneuvers such as continuing straight and turning at intersections, as well as following traffic and stopping. Section~\ref{sec:joint_pretraining} describes the joint pretraining setup, including the trainable modules and the four loss terms.

\noindent\textbf{NAVSIM fine-tuning and post-training.}
We fine-tune \model~on four-frame clips from the NAVSIM \emph{navtrain} split~\cite{bench1,bench2} using its native navigation inputs. We report the supervised checkpoint and a second checkpoint obtained by applying Flow-GRPO to its trajectory planner on the same split. Section~\ref{sec:navsim_grpo} describes the reward and update, and Tab.~\ref{tab:flow_grpo_config} gives the post-training configuration.

\noindent\textbf{NAVSIM benchmarks and metrics.}
We evaluate on NAVSIM v1 \emph{navtest}~\cite{bench1} and on NAVSIM v2 \emph{navtest} and \emph{navhard}~\cite{bench2}. For v1, we report PDMS together with no-at-fault collision (NC), drivable area compliance (DAC), ego progress (EP), time-to-collision (TTC), and comfort (C.). For v2, we report EPDMS and its component scores, including driving-direction compliance (DDC), traffic-light compliance (TLC), lane keeping (LK), history comfort (HC), and extended comfort (EC). Higher scores indicate better performance. On v2 \emph{navtest}, we use single-stage EPDMS with the human penalty. On \emph{navhard}, the protocol combines scores from two pseudo-closed-loop stages with reactive traffic. For every split, we compute all metrics from the single trajectory that the planner commits to for each scene.

\noindent\textbf{nuScenes.}
For zero-shot transfer, we apply the NAVSIM-trained model to the nuScenes validation split, which contains 150 of the dataset's 1,000 scenes~\cite{caesar2020nuscenes}. Tab.~\ref{tab:zero_shot_transfer} reports L2 displacement error in meters and collision rate in percent at 1, 2, and 3~s, along with the average for each metric.

\noindent\textbf{Bench2Drive.}
Bench2Drive~\cite{jia2024bench2drive} is a CARLA v2 closed-loop benchmark~\cite{dosovitskiy2017carla} with interactive scenarios and evaluation routes. To assess transfer from real driving data to simulation, we test the NAVSIM-trained model on the Bench2Drive validation split without target-domain fine-tuning. We use the same displacement and collision metrics and evaluation horizons as for nuScenes.

\noindent\textbf{Model configuration.}
We initialize the world-model backbone with the language transformer of Qwen3-VL 2B~\cite{bai2025qwen3vl}. \model~contains 3.1B parameters in total, of which 2.4B are retained for evaluation and real-vehicle deployment. At inference, the planner reads scene features from the shared backbone and decodes its trajectory using the pathway described in Sec.~\ref{sec:method_trajae_diffusion}, with the auxiliary heads disabled.

\noindent\textbf{NAVSIM implementation.}
For NAVSIM evaluation, the model receives a history of four frames sampled at 2\,Hz, with input images at \(512 \times 1024\) resolution. We run inference on a single NVIDIA RTX~4090 GPU and evaluate the single trajectory produced from each four-frame observation history.

\subsection{NAVSIM Results}

Tab.~\ref{tab:navtest-v1}, Tab.~\ref{tab:navtest-v2}, and Tab.~\ref{tab:navhard-v2} report the aggregate and component scores for the three NAVSIM splits.
\begin{table}[!htbp]
\centering
  \caption[NAVSIM v1 (navtest)]{\textbf{NAVSIM v1 on navtest}~\cite{bench1}. PDMS and its components score the single trajectory selected by each planner, with higher values indicating better performance across all five criteria.}
  \label{tab:navtest-v1}
  \small
\newcommand{\NavsimCamera}[2]{$#1$C#2}
\newcommand{\NavsimOneRows}{%
  \NavsimOneRow{Human}{- & - & $100.0$}{$100.0$ & $87.5$ & $100.0$ & $99.9$}{$94.8$}%
  \midrule%
  \NavsimOneRow{TransFuser~\cite{chitta2022transfuser}}{\NavsimCamera{3}{+L} & \xmark & $97.7$}{$92.8$ & $79.2$ & $92.8$ & $\mathbf{100.0}$}{$84.0$}%
  \NavsimOneRow{Epona~\cite{zhang2025epona}}{\NavsimCamera{1}{} & \xmark & $97.9$}{$95.1$ & $80.4$ & $93.8$ & $99.9$}{$86.2$}%
  \NavsimOneRow{DiffusionDrive~\cite{liao2025diffusiondrive}}{\NavsimCamera{3}{+L} & \xmark & $98.2$}{$96.2$ & $82.2$ & $94.7$ & $\mathbf{100.0}$}{$88.1$}%
  \NavsimOneRow{VGGDrive~\cite{wang2026vggdrive}}{\NavsimCamera{3}{} & \xmark & $98.6$}{$96.3$ & $82.9$ & $95.6$ & $\mathbf{100.0}$}{$88.8$}%
  \NavsimOneRow{AutoVLA~\cite{zhou2025autovla}}{\NavsimCamera{3}{} & \xmark & $96.9$}{$92.4$ & $75.8$ & $88.1$ & $99.9$}{$80.5$}%
  \NavsimOneRow{DriveLaW~\cite{xia2026drivelaw}}{\NavsimCamera{1}{} & \xmark & $99.0$}{$97.1$ & $81.3$ & $96.7$ & $\mathbf{100.0}$}{$89.1$}%
  \NavsimOneRow{EponaV2~\cite{xu2026eponav2}}{\NavsimCamera{1}{} & \xmark & $98.6$}{$97.3$ & $83.6$ & $95.3$ & $99.9$}{$89.4$}%
  \NavsimOneRow{DriveVLA-W0~\cite{li2025drivevlaw0}}{\NavsimCamera{1}{} & \xmark & $98.7$}{$\mathbf{99.1}$ & $83.3$ & $95.3$ & $99.3$}{$90.2$}%
  \NavsimOneRow{HyWorldVLA~\cite{yu2026hyworldvla}}{\NavsimCamera{1}{} & \xmark & $98.6$}{$98.9$ & $84.6$ & $95.1$ & $99.9$}{$90.6$}%
  \NavsimOneRow{DriveVA~\cite{liu2026driveva}}{\NavsimCamera{1}{} & \xmark & $99.2$}{$97.5$ & $83.5$ & $\mathbf{98.7}$ & $\mathbf{100.0}$}{$90.9$}%
  \NavsimOneRow{Auto-JEPA~\cite{yang2026auto}}{\NavsimCamera{1}{} & \xmark & $98.4$}{$98.3$ & $87.1$ & $95.0$ & $\mathbf{100.0}$}{$91.3$}%
  \NavsimOneRow{PhysWAM~\cite{physwam}}{\NavsimCamera{3}{} & \xmark & $99.0$}{$97.5$ & $\mathbf{88.7}$ & $98.5$ & $99.8$}{$91.4$}%
  \midrule%
  \NavsimOneRow{\model~(Ours)}{\NavsimCamera{6}{} & \xmark & $\mathbf{99.3}$}{$98.5$ & $85.0$ & $97.4$ & $\mathbf{100.0}$}{$\mathbf{91.9}$}%
  \midrule%
  \NavsimOneRow{AutoVLA~\cite{zhou2025autovla}}{\NavsimCamera{3}{} & \cmark & $98.4$}{$95.6$ & $81.9$ & $98.0$ & $99.9$}{$89.1$}%
  \NavsimOneRow{AutoDrive-P$^3$~\cite{anonymous2026autodrivep3}}{\NavsimCamera{3}{} & \cmark & $99.1$}{$97.4$ & $84.8$ & $96.5$ & $\mathbf{100.0}$}{$90.6$}%
  \NavsimOneRow{ReCogDrive~\cite{li2025recogdrive}}{\NavsimCamera{1}{} & \cmark & $97.9$}{$97.3$ & $\mathbf{87.3}$ & $94.9$ & $\mathbf{100.0}$}{$90.8$}%
  \NavsimOneRow{ExploreVLA~\cite{sheng2026explorevla}}{\NavsimCamera{1}{} & \cmark & $98.8$}{$98.4$ & $83.5$ & $96.5$ & $99.9$}{$90.4$}%
  \NavsimOneRow{EponaV2~\cite{xu2026eponav2}}{\NavsimCamera{1}{} & \cmark & $98.6$}{$97.9$ & $84.8$ & $95.7$ & $\mathbf{100.0}$}{$90.4$}%
  \NavsimOneRow{Qwen-Drive-RL~\cite{zhou2026qwen}}{\NavsimCamera{3}{} & \cmark & $98.6$}{$98.2$ & $84.8$ & $95.9$ & $\mathbf{100.0}$}{$90.7$}%
  \midrule%
  \NavsimOneRow{\model~(Ours)}{\NavsimCamera{6}{} & \cmark & $\mathbf{99.3}$}{$\mathbf{98.9}$ & $85.7$ & $\mathbf{97.6}$ & $\mathbf{100.0}$}{$\mathbf{92.5}$}%
}

\NavsimPrepareScores{navsimScoreBlue}{\NavsimOneRows}{\NavsimOneRow}
\renewcommand{\NavsimOneRow}[4]{#1 & #2 & #3 & \NavsimShadeScore{#1}{#4}#4\\}
\begin{tabular*}{\textwidth}{@{\extracolsep{\fill}}lccccccc>{\columncolor{white}[\tabcolsep][0pt]}c@{}}
\toprule
\textbf{Model} & \textbf{Input} & \textbf{RL} & \textbf{NC} & \textbf{DAC} & \textbf{EP} & \textbf{TTC} & \textbf{C.} & \textbf{PDMS} \\
\midrule
\NavsimOneRows
\bottomrule
\end{tabular*}
\end{table}

\begin{table}[!htbp]
\centering
  \caption[NAVSIM v2 (navtest)]{\textbf{NAVSIM v2 navtest}~\cite{bench2}. Single-stage EPDMS includes the human penalty. The table groups methods by whether they use RL post-training for trajectory planning.}
  \label{tab:navtest-v2}
  \small
  \setlength{\tabcolsep}{2.5pt}
\newcommand{\NavsimTwoRows}{%
  \NavsimTwoRow{Human}{- & $100.0$ & $100.0$}{\NavsimTwoMetrics{$99.8$ & $100.0$ & $97.8$ & $100.0$ & $100.0$ & $98.1$}{$90.1$}}{$97.9$}%
  \midrule%
  \NavsimTwoRow{TransFuser~\cite{chitta2022transfuser}}{\xmark & $96.9$ & $89.9$}{\NavsimTwoMetrics{$97.8$ & $99.7$ & $87.1$ & $95.4$ & $92.7$ & $98.3$}{$87.2$}}{$84.0$}%
  \NavsimTwoRow{Epona~\cite{zhang2025epona}}{\xmark & $97.9$ & $95.2$}{\NavsimTwoMetrics{$99.4$ & $99.8$ & $86.5$ & $97.1$ & $96.5$ & $98.3$}{$70.4$}}{$85.2$}%
  \NavsimTwoRow{DriveVLA-W0~\cite{li2025drivevlaw0}}{\xmark & $98.4$ & $95.2$}{\NavsimTwoMetrics{$99.4$ & $\mathbf{99.9}$ & $86.6$ & $97.9$ & $97.8$ & $98.3$}{$82.7$}}{$86.9$}%
  \NavsimTwoRow{WoTE~\cite{li2025wote}}{\xmark & $98.5$ & $96.8$}{\NavsimTwoMetrics{$98.8$ & $99.8$ & $86.1$ & $97.9$ & $95.5$ & $98.3$}{$82.9$}}{$87.7$}%
  \NavsimTwoRow{DiffusionDrive~\cite{liao2025diffusiondrive}}{\xmark & $98.2$ & $96.2$}{\NavsimTwoMetrics{$99.5$ & $99.8$ & $87.4$ & $97.3$ & $96.9$ & $98.4$}{$87.7$}}{$88.2$}%
  \NavsimTwoRow{PWM~\cite{zhao2025pwm}}{\xmark & $98.8$ & $95.9$}{\NavsimTwoMetrics{$99.4$ & $\mathbf{99.9}$ & $86.4$ & $98.4$ & $97.6$ & $98.3$}{$85.3$}}{$88.2$}%
  \NavsimTwoRow{DriveLaW~\cite{xia2026drivelaw}}{\xmark & $98.7$ & $96.9$}{\NavsimTwoMetrics{$99.6$ & $99.8$ & $87.5$ & $98.3$ & $97.6$ & $98.4$}{$77.4$}}{$88.6$}%
  \NavsimTwoRow{Auto-JEPA~\cite{yang2026auto}}{\xmark & $98.5$ & $\mathbf{98.7}$}{\NavsimTwoMetrics{$98.3$ & $99.7$ & $\mathbf{90.5}$ & $97.9$ & $94.7$ & $97.8$}{$75.2$}}{$89.1$}%
  \NavsimTwoRow{HyWorldVLA~\cite{yu2026hyworldvla}}{\xmark & $98.8$ & $98.2$}{\NavsimTwoMetrics{$99.5$ & $\mathbf{99.9}$ & $87.8$ & $98.1$ & $96.2$ & $98.3$}{$77.4$}}{$89.7$}%
  \NavsimTwoRow{PhysWAM~\cite{physwam}}{\xmark & $99.0$ & $97.5$}{\NavsimTwoMetrics{$99.3$ & $99.8$ & $88.7$ & $98.5$ & $96.5$ & $\mathbf{98.6}$}{$\mathbf{90.5}$}}{$90.3$}%
  \midrule%
  \NavsimTwoRow{\model~(Ours)}{\xmark & $\mathbf{99.2}$ & $98.4$}{\NavsimTwoMetrics{$\mathbf{99.8}$ & $\mathbf{99.9}$ & $87.4$ & $\mathbf{98.9}$ & $\mathbf{98.5}$ & $98.3$}{$81.4$}}{$\mathbf{91.1}$}%
  \midrule%
  \NavsimTwoRow{ExploreVLA~\cite{sheng2026explorevla}}{\cmark & $98.8$ & $96.2$}{\NavsimTwoMetrics{$99.6$ & $99.8$ & $87.1$ & $98.2$ & $97.8$ & $98.3$}{$\mathbf{86.8}$}}{$88.8$}%
  \NavsimTwoRow{EponaV2~\cite{xu2026eponav2}}{\cmark & $98.5$ & $97.4$}{\NavsimTwoMetrics{$99.5$ & $\mathbf{99.9}$ & $87.9$ & $98.1$ & $97.7$ & $98.2$}{$77.4$}}{$88.9$}%
  \NavsimTwoRow{AutoDrive-P$^3$~\cite{anonymous2026autodrivep3}}{\cmark & $99.1$ & $97.4$}{\NavsimTwoMetrics{$99.2$ & $99.8$ & $\mathbf{88.0}$ & $98.7$ & $96.3$ & $98.3$}{$85.5$}}{$89.9$}%
  \midrule%
  \NavsimTwoRow{\model~(Ours)}{\cmark & $\mathbf{99.3}$ & $\mathbf{98.6}$}{\NavsimTwoMetrics{$\mathbf{99.8}$ & $\mathbf{99.9}$ & $\mathbf{88.0}$ & $\mathbf{99.0}$ & $\mathbf{98.6}$ & $\mathbf{98.4}$}{$83.1$}}{$\mathbf{91.8}$}%
}

\NavsimPrepareScores{navsimScorePurple}{\NavsimTwoRows}{\NavsimTwoRow}
\newcommand{\NavsimTwoMetrics}[2]{#1 & #2}
\renewcommand{\NavsimTwoRow}[4]{#1 & #2 & #3 & \NavsimShadeScore{#1}{#4}#4\\}
\begin{tabular*}{\textwidth}{@{\extracolsep{\fill}}lcccccccccc>{\columncolor{white}[\tabcolsep][0pt]}c@{}}
\toprule
\textbf{Model} & \textbf{RL} & \textbf{NC} & \textbf{DAC} & \textbf{DDC} & \textbf{TLC} & \textbf{EP} & \textbf{TTC} & \textbf{LK} & \textbf{HC} & \textbf{EC} & \textbf{EPDMS} \\
\midrule
\NavsimTwoRows
\bottomrule
\end{tabular*}
\end{table}

\begin{table}[!htbp]
\centering
  \caption[NAVSIM v2 (navhard)]{\textbf{NAVSIM v2 navhard}~\cite{bench2}. Combined EPDMS aggregates scores from two stages of pseudo-simulation with reactive traffic. The component scores are shown separately for each stage.}
  \label{tab:navhard-v2}
  \small
  \setlength{\tabcolsep}{2.5pt}
\newcommand{\NavsimHardMetrics}[9]{#1 & #2 & #3 & #4 & #5 & #6 & #7 & #8 & #9}
\newcommand{\NavsimHardRows}{%
  \NavsimHardRow{DriveVLA-W0~\cite{li2025drivevlaw0}}%
    {\NavsimHardMetrics{$96.8$}{$83.3$}{$99.0$}{$99.6$}{$84.6$}{$95.3$}{$96.4$}{$97.6$}{$78.2$}}%
    {\NavsimHardMetrics{$76.8$}{$64.3$}{$79.9$}{$98.3$}{$\mathbf{89.2}$}{$75.0$}{$46.8$}{$95.8$}{$53.1$}}%
    {$24.4$}%
  \midrule%
  \NavsimHardRow{ReCogDrive~\cite{li2025recogdrive}}%
    {\NavsimHardMetrics{$96.4$}{$78.9$}{$98.7$}{$99.8$}{$82.6$}{$95.6$}{$94.4$}{$97.6$}{$74.2$}}%
    {\NavsimHardMetrics{$80.2$}{$65.0$}{$82.4$}{$98.7$}{$85.2$}{$76.9$}{$43.8$}{$\mathbf{96.6}$}{$71.8$}}%
    {$25.7$}%
  \midrule%
  \NavsimHardRow{DiffusionDrive~\cite{liao2025diffusiondrive}}%
    {\NavsimHardMetrics{$96.8$}{$86.0$}{$98.8$}{$99.3$}{$84.0$}{$95.8$}{$96.7$}{$97.6$}{$\mathbf{79.6}$}}%
    {\NavsimHardMetrics{$80.1$}{$72.8$}{$84.4$}{$98.4$}{$85.9$}{$76.6$}{$46.4$}{$96.3$}{$\mathbf{72.8}$}}%
    {$27.5$}%
  \midrule%
  \NavsimHardRow{DriveLaW~\cite{xia2026drivelaw}}%
    {\NavsimHardMetrics{$97.3$}{$89.1$}{$99.2$}{$99.6$}{$84.3$}{$97.1$}{$96.2$}{$\mathbf{97.8}$}{$67.6$}}%
    {\NavsimHardMetrics{$82.5$}{$67.6$}{$83.5$}{$98.1$}{$84.8$}{$78.5$}{$45.8$}{$96.4$}{$57.3$}}%
    {$30.6$}%
  \midrule%
  \NavsimHardRow{GeoWAM~\cite{geowam}}%
    {\NavsimHardMetrics{$97.7$}{$91.5$}{$99.1$}{$99.8$}{$83.8$}{$95.8$}{$96.0$}{$\mathbf{97.8}$}{$79.0$}}%
    {\NavsimHardMetrics{$80.4$}{$76.3$}{$87.3$}{$98.7$}{$88.9$}{$76.2$}{$49.9$}{$94.0$}{$56.0$}}%
    {$36.6$}%
  \midrule%
  \NavsimHardRow{PhysWAM~\cite{physwam}}%
    {\NavsimHardMetrics{$96.9$}{$90.9$}{$99.0$}{$99.6$}{$83.8$}{$95.6$}{$95.8$}{$\mathbf{97.8}$}{$67.6$}}%
    {\NavsimHardMetrics{$83.9$}{$\mathbf{80.2}$}{$87.3$}{$98.7$}{$88.7$}{$81.6$}{$48.9$}{$95.3$}{$54.4$}}%
    {$38.1$}%
  \midrule%
  \NavsimHardRow{Epona~\cite{zhang2025epona}}%
    {\NavsimHardMetrics{$96.7$}{$88.9$}{$98.1$}{$99.6$}{$84.0$}{$95.6$}{$94.4$}{$97.6$}{$58.2$}}%
    {\NavsimHardMetrics{$80.7$}{$74.5$}{$86.6$}{$98.7$}{$87.0$}{$77.2$}{$49.7$}{$96.5$}{$50.2$}}%
    {$31.6$}%
  \midrule%
  \NavsimHardRow{EponaV2~\cite{xu2026eponav2}}%
    {\NavsimHardMetrics{$97.3$}{$90.7$}{$99.4$}{$\mathbf{100.0}$}{$83.3$}{$97.3$}{$97.3$}{$97.6$}{$60.9$}}%
    {\NavsimHardMetrics{$83.6$}{$78.0$}{$88.0$}{$\mathbf{98.9}$}{$86.0$}{$80.3$}{$50.1$}{$96.1$}{$52.0$}}%
    {$36.1$}%
  \midrule%
  \NavsimHardRow{\model~(Ours)}%
    {\NavsimHardMetrics{$\mathbf{98.4}$}{$\mathbf{93.6}$}{$\mathbf{99.6}$}{$\mathbf{100.0}$}{$\mathbf{84.8}$}{$\mathbf{98.4}$}{$\mathbf{97.8}$}{$\mathbf{97.8}$}{$72.9$}}%
    {\NavsimHardMetrics{$\mathbf{85.9}$}{$79.6$}{$\mathbf{89.2}$}{$98.7$}{$85.7$}{$\mathbf{81.8}$}{$\mathbf{50.7}$}{$96.4$}{$65.3$}}%
    {$\mathbf{48.3}$}%
}

\NavsimPrepareScores{navsimScoreBlue}{\NavsimHardRows}{\NavsimHardRow}
\renewcommand{\NavsimHardRow}[4]{%
  & $1$ & #2 & \NavsimShadeScore{#1}{#4}\\
  \multirow{-2}{*}{#1} & $2$ & #3 & \NavsimShadeScore{#1}{#4}\multirow{-2}{*}{#4}\\}
\begin{tabular*}{\textwidth}{@{\extracolsep{\fill}}lcccccccccc>{\columncolor{white}[\tabcolsep][0pt]}c@{}}
\toprule
\textbf{Model} & \textbf{Stage} & \textbf{NC} & \textbf{DAC} & \textbf{DDC} & \textbf{TLC} & \textbf{EP} & \textbf{TTC} & \textbf{LK} & \textbf{HC} & \textbf{EC} & \textbf{EPDMS} \\
\midrule
\NavsimHardRows
\bottomrule
\end{tabular*}
{\footnotesize\raggedright\noindent
``C'' counts camera views, and ``L'' denotes LiDAR. NAVSIM v2 uses the human penalty. Gray marks the human row, and darker shading indicates higher model scores on the same metric.\\
Bold marks the best scores in each RL group on navtest and across methods on navhard.\par}
\end{table}

\noindent\textbf{NAVSIM v1 navtest.}
As shown in Tab.~\ref{tab:navtest-v1}, supervised fine-tuning reaches \textbf{91.9 PDMS} with one trajectory per scene. This is 0.5 points above PhysWAM~\cite{physwam}, the highest-scoring non-RL baseline in the table, and 1.2 points above Qwen-Drive-RL~\cite{zhou2026qwen}. Flow-GRPO adds 0.6 points to reach \textbf{92.5 PDMS}, 2.1 points above EponaV2~\cite{xu2026eponav2} under the same protocol. The supervised checkpoint generates its trajectory directly, and the second checkpoint uses the same planner after reward-based post-training.

\noindent\textbf{NAVSIM v2 navtest.}
Tab.~\ref{tab:navtest-v2} shows that \model~reaches \textbf{91.1 EPDMS} before GRPO and \textbf{91.8 EPDMS} afterward, 1.5 points above the single-sample result of PhysWAM~\cite{physwam}. The model improves on the 6.7B-parameter EponaV2~\cite{xu2026eponav2} by 2.9 points with 3.1B parameters. Among the RL-based methods in the table, \model~matches or exceeds the highest score on eight of the nine component metrics.

\noindent\textbf{NAVSIM v2 navhard.}
On navhard, \model~achieves a \textbf{combined EPDMS of 48.3} (Tab.~\ref{tab:navhard-v2}), outperforming PhysWAM~\cite{physwam} by 10.2 points and EponaV2~\cite{xu2026eponav2} by 12.2 points. In Stage~1, it leads or ties on eight of the nine component metrics. In Stage~2, it ranks first among the reported methods on no-at-fault collision, driving-direction compliance, time-to-collision, and lane keeping under reactive traffic. These results show that \model~performs particularly well on closed-loop driving metrics, consistent with its strong performance in real-world driving tests.

\begin{figure}[!htp]
    \centering
    \includegraphics[width=\textwidth]{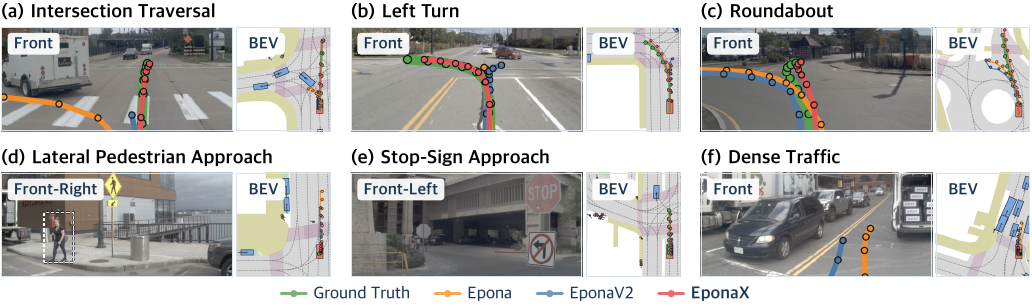}
    \caption{\textbf{Qualitative planning on NAVSIM navtest~\cite{bench1}.} Each camera image is paired with a bird's-eye view showing the predictions of Epona~\cite{zhang2025epona}, EponaV2~\cite{xu2026eponav2}, and \model~alongside the recorded trajectory. Fig.~\ref{fig:series_trajectory_cases} gives further examples of turn direction, curvature, and progress.}
    \label{fig:qualitative_planning}
\end{figure}

\noindent\textbf{Qualitative planning results.}
Fig.~\ref{fig:qualitative_planning} compares planned trajectories in six navtest scenarios. \model~closely follows the recorded trajectory when crossing an intersection in panel (a) and turning left in panel (b). Panel (d) marks a pedestrian approaching from the right, and the front-left view in panel (e) shows the stop sign. Additional comparisons in Fig.~\ref{fig:series_trajectory_cases} pair camera views with planned paths to compare how the models choose turn direction, curvature, and progress along the route in the selected scenarios.

\subsection{Cross-Domain Generalization}

\begin{table}[!htp]
\centering
\captionsetup{singlelinecheck=false,justification=raggedright}
\setlength{\tabcolsep}{1.75pt}
\caption{\textbf{Zero-shot cross-domain motion planning.} Methods trained on NAVSIM~\cite{bench1} are evaluated on nuScenes~\cite{caesar2020nuscenes} and Bench2Drive (CARLA)~\cite{jia2024bench2drive} without target-domain fine-tuning.
L2 error and collision rate are reported at 1, 2, and 3~s and averaged over these horizons.}
\label{tab:zero_shot_transfer}
\fontsize{7.6}{9.4}\selectfont
\begin{tabular*}{\linewidth}{@{\extracolsep{\fill}}l*{2}{ccc>{\columncolor{navsimScoreBlue!10}}c}*{2}{ccc>{\columncolor{navsimScorePurple!10}}c}@{}}
\toprule
& \multicolumn{8}{c}{\textbf{nuScenes}}
& \multicolumn{8}{c}{\textbf{Bench2Drive (CARLA)}} \\
\cmidrule(lr){2-9}
\cmidrule(lr){10-17}
& \multicolumn{4}{c}{L2 (m) $\downarrow$}
& \multicolumn{4}{c}{Collision Rate (\%) $\downarrow$}
& \multicolumn{4}{c}{L2 (m) $\downarrow$}
& \multicolumn{4}{c}{Collision Rate (\%) $\downarrow$} \\
\cmidrule(lr){2-5}
\cmidrule(lr){6-9}
\cmidrule(lr){10-13}
\cmidrule(lr){14-17}
\textbf{Method}
& \textbf{1s} & \textbf{2s} & \textbf{3s} & \cellcolor{white}\textbf{Avg.}
& \textbf{1s} & \textbf{2s} & \textbf{3s} & \cellcolor{white}\textbf{Avg.}
& \textbf{1s} & \textbf{2s} & \textbf{3s} & \cellcolor{white}\textbf{Avg.}
& \textbf{1s} & \textbf{2s} & \textbf{3s} & \cellcolor{white}\textbf{Avg.} \\
\midrule
PWM~\cite{zhao2025pwm}
& $2.06$ & $3.91$ & $6.00$ & $3.99$
& $0.12$ & $0.15$ & $0.86$ & $0.36$
& $1.70$ & $2.74$ & $3.97$ & $2.80$
& $4.01$ & $3.73$ & $3.53$ & $3.76$ \\
DriveVLA-W0~\cite{li2025drivevlaw0}
& $0.43$ & $1.26$ & $2.60$ & $1.43$
& $0.22$ & $0.66$ & $1.42$ & $0.77$
& $1.01$ & $2.77$ & $5.22$ & $3.00$
& $1.49$ & $2.53$ & $3.53$ & $2.52$ \\
DriveVA~\cite{liu2026driveva}
& $\mathbf{0.33}$ & $\mathbf{0.76}$ & $1.43$ & $\mathbf{0.84}$
& $\mathbf{0.00}$ & $0.07$ & $0.12$ & $0.06$
& $0.69$ & $1.29$ & $2.03$ & $1.33$
& $1.38$ & $1.97$ & $2.65$ & $1.79$ \\
\midrule
\model~(Ours)
& $0.47$ & $0.81$ & $\mathbf{1.26}$ & $0.85$
& $0.02$ & $\mathbf{0.02}$ & $\mathbf{0.10}$ & $\mathbf{0.05}$
& $\mathbf{0.57}$ & $\mathbf{1.13}$ & $\mathbf{1.83}$ & $\mathbf{1.18}$
& $\mathbf{1.09}$ & $\mathbf{1.19}$ & $\mathbf{1.69}$ & $\mathbf{1.33}$ \\
\bottomrule
\end{tabular*}
\end{table}

Tab.~\ref{tab:zero_shot_transfer} compares methods trained on NAVSIM~\cite{bench1} and evaluated directly on nuScenes~\cite{caesar2020nuscenes} and Bench2Drive~\cite{jia2024bench2drive}. On nuScenes, \model~achieves an average L2 error of 0.85~m, close to DriveVA's 0.84~m, and the lowest average collision rate at \textbf{0.05\%}. Relative to PWM, the average L2 error drops from 3.99~m and the collision rate from 0.36\%, narrowing both errors under the same transfer setup.

On Bench2Drive, \model~achieves the lowest L2 error and collision rate at each of the three evaluation horizons under real-to-simulation transfer. It obtains an average L2 error of \textbf{1.18~m} and an average collision rate of \textbf{1.33\%}, improving over DriveVA by 11.3\% and 25.7\%, respectively. The planner achieves these gains using the same NAVSIM training and evaluation horizons as in the nuScenes transfer.

\subsection{Real-World Deployment Evaluation}
\label{sec:real_vehicle_deployment}

\noindent\textbf{Onboard implementation.}
The deployed policy uses three historical observations from six cameras and structured navigation, as defined in Sec.~\ref{sec:method_overview}. Fig.~\ref{fig:fig_camera_2} locates the front-wide camera, whose view is tokenized at higher spatial detail, and the five calibrated surround cameras, whose views are compressed jointly. Two parallel inference replicas run on two NVIDIA RTX~5090 GPUs and sustain a 10\,Hz control update rate through visual tokenization, backbone processing, and trajectory decoding.

\noindent\textbf{Route protocol.}
We drive three laps on each of four urban loops in Shanghai, covering approximately 240\,km. Tab.~\ref{tab:route-locations} gives route lengths and street names, and Fig.~\ref{fig:roads_appendix} maps the loops. Testing covers daytime and nighttime driving, rain, highway segments, and strong illumination. We record each driver intervention together with the event that prompted it during the twelve evaluation laps.

\begingroup
\setlength{\intextsep}{9pt plus 1pt minus 1pt}
\begin{figure}[!htp]
    \centering
    \begin{minipage}[t]{0.55\linewidth}
        \vspace{0pt}\centering
        \includegraphics[width=\linewidth]{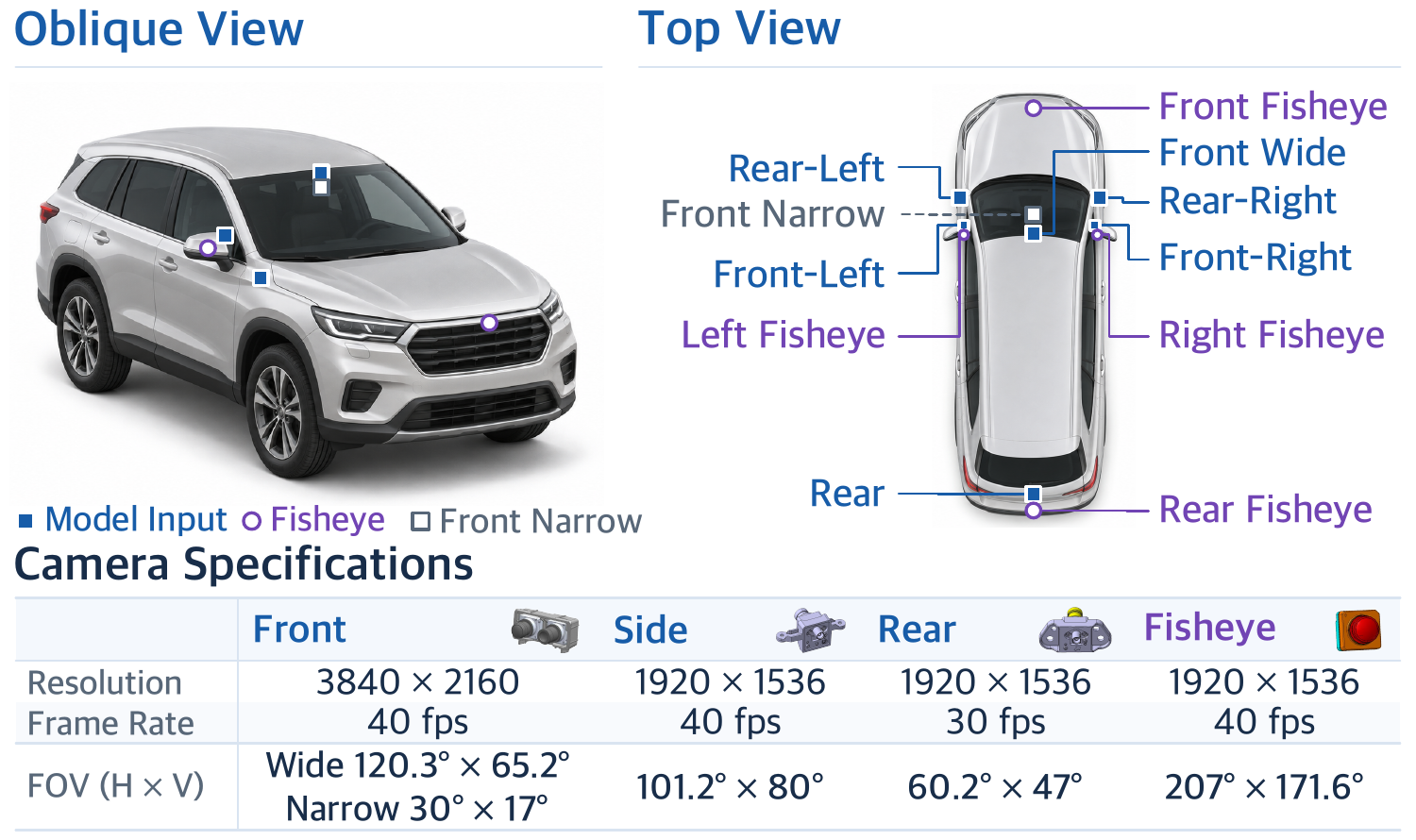}
        \captionsetup{font=small}
        \caption{\textbf{Onboard camera layout.} Filled squares mark the six model inputs. Circles indicate fisheye cameras, and the outlined square marks the narrow front camera on the vehicle.}
        \label{fig:fig_camera_2}
    \end{minipage}\hfill
    \begin{minipage}[t]{0.425\linewidth}
        \vspace{0pt}\centering
        \includegraphics[width=\linewidth]{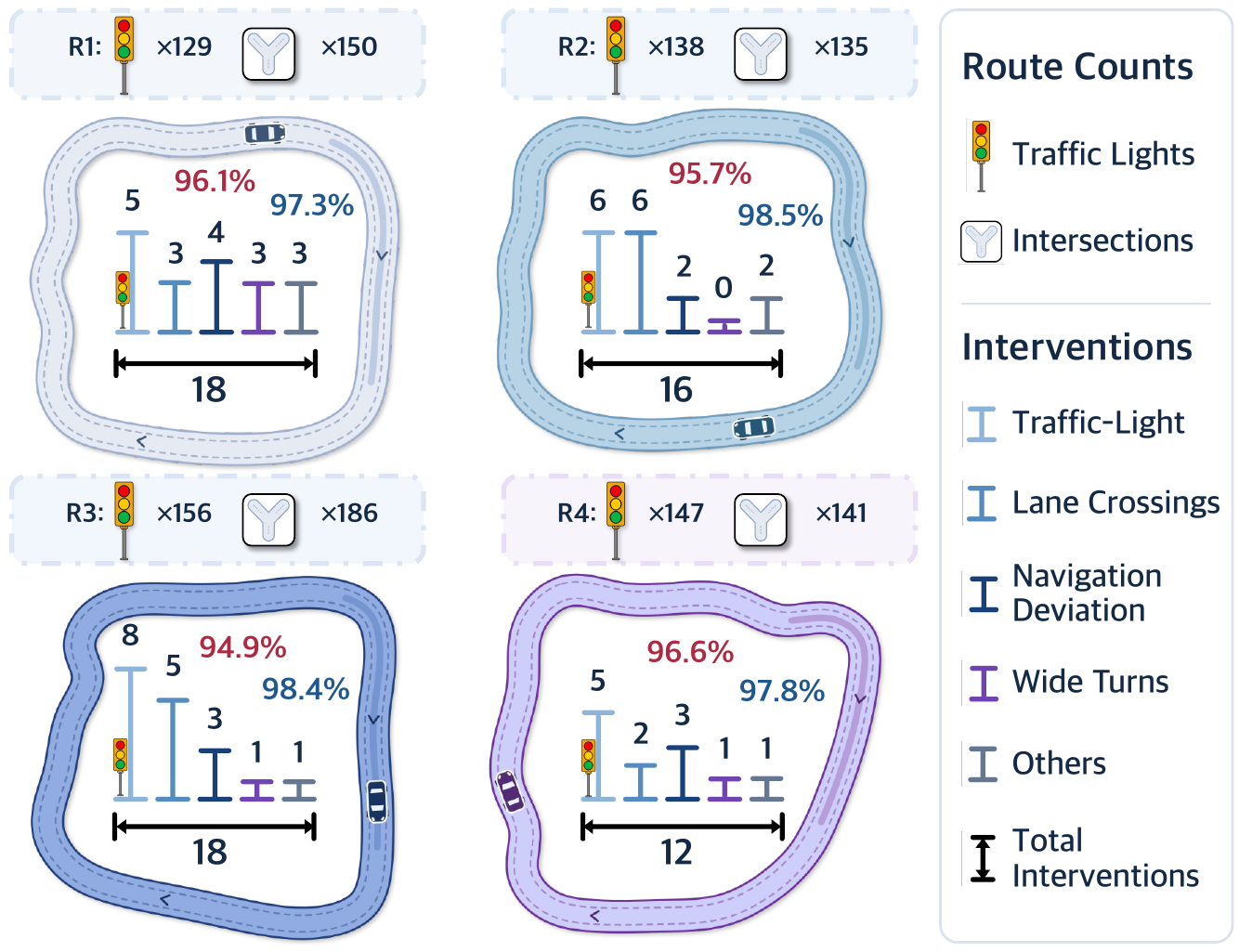}
        \captionsetup{font=small}
        \caption{\textbf{Interventions by route.} Counts cover three laps per loop. Red marks traffic-light compliance and blue marks navigation compliance.}
        \label{fig:Table_5}
    \end{minipage}
\end{figure}

\begin{trainingconfiguration}
    {\textbf{Real-vehicle evaluation routes in Shanghai.} The lengths are estimates for a single lap.
    We drove three laps per route, and Fig.~\ref{fig:roads_appendix} shows the detailed maps of the four loops.}
    {tab:route-locations}
    \begin{tabularx}{\linewidth}{@{}>{\raggedright\arraybackslash}X@{\hspace{20pt}}>{\raggedright\arraybackslash}X@{}}
        \reporthead{Test Loops} & \reporthead{Selected Streets} \\
        \noalign{\vskip 1pt{\color{reportAccent!24!white}\hrule height 0.4pt}\vskip 2pt}
        \makebox[17pt][l]{\textbf{R1}}19.47\,km\hspace{8pt}Zhangjiang Innovation Park
            & Longdong Ave., Jinke Rd., Gaoke Middle Rd. \\
        \addlinespace[2pt]
        \makebox[17pt][l]{\textbf{R2}}19.35\,km\hspace{8pt}Zhangjiang East and Sunqiao
            & Zhangjiang Rd., Zhangdong Rd., Sunjian Rd. \\
        \addlinespace[2pt]
        \makebox[17pt][l]{\textbf{R3}}19.63\,km\hspace{8pt}Chuansha
            & Huaxia East Rd., Chuansha Rd., Chuanzhou Hwy. \\
        \addlinespace[2pt]
        \makebox[17pt][l]{\textbf{R4}}20.33\,km\hspace{8pt}Caotanghe
            & Huadong Rd., Dongchuan Hwy., Longdong Ave. \\
    \end{tabularx}
\end{trainingconfiguration}
\endgroup

\noindent\textbf{Intervention analysis.}
Fig.~\ref{fig:Table_5} shows 18, 16, 18, and 12 interventions on R1 through R4, for a total of \textbf{64}. Rule-related issues account for approximately 80\% of interventions, chiefly traffic-light responses and navigation deviations such as choosing a lane inconsistent with route guidance. Drivers intervened at 24 of 570 traffic-light encounters, corresponding to 4.2\%. The policy followed navigation guidance in approximately \textbf{98.0\%} of cases across the four routes during the deployment evaluation.

Across all twelve laps, wide turns and other safety-related causes account for 12 interventions over approximately 240\,km, yielding a pooled rate of about \textbf{5} interventions per 100\,km. Based on the estimated distance driven over three laps on each route, three of the four routes fall within the range of \textbf{3--5} interventions per 100\,km, with rates of 3.4 on both R2 and R3 and 3.3 on R4. R1 has the highest rate at 10.3 interventions per 100\,km. Section~\ref{sec:benchmark_vehicle_ablations} reports interventions on a single lap of R1 for each ablation variant.

\subsection{Benchmark and Real-Vehicle Ablation Studies}
\label{sec:benchmark_vehicle_ablations}

\noindent\textbf{Real-vehicle ablation protocol.}
We test each variant on one lap of the approximately 20\,km R1 route using the same vehicle, controller, and safety-driver procedure. Traffic varies by lap. The driver assesses start-up success at opportunities to move off, such as when a lead vehicle starts or a light turns green. A successful start is prompt and appropriate to the traffic situation and requires no driver intervention.

Braking and avoidance success is measured in encounters requiring deceleration, stopping, or evasive action. A successful response is timely and avoids a collision or driver intervention. The intervention count includes manual braking, steering, and takeovers to maintain safe operation throughout the lap.

\newsavebox{\ablationInputTable}
\begin{table}[!htp]
\centering
\captionsetup{font=small,justification=raggedright,skip=5pt}

\begin{lrbox}{\ablationInputTable}
\begin{minipage}[b]{0.51\linewidth}
\caption{\textbf{Input and supervision ablations.} PDMS and EPDMS use NAVSIM v1 navtest and v2 navhard, respectively. Each vehicle test covers one lap of R1. Dashes indicate unreported start-up and braking results.}
\label{tab:ablation-input-supervision}
\footnotesize
\setlength{\tabcolsep}{1.15pt}
\renewcommand{\arraystretch}{1.22}
\par\nointerlineskip
\hbox{\begin{tabular*}{\linewidth}{@{\extracolsep{\fill}}lcccc>{\columncolor{navsimScorePurple!10}[\tabcolsep][0pt]}c@{}}
\toprule
& \multicolumn{2}{c}{\textbf{Simulation}} & \multicolumn{3}{c}{\textbf{Real Vehicle}} \\
\cmidrule(lr){2-3} \cmidrule(l){4-6}
\textbf{Setting} & \textbf{PDMS} & \textbf{EPDMS}
    & \shortstack{\textbf{Start-Up}\\\textbf{(\%)}}
    & \shortstack{\textbf{Brake/}\\\textbf{Avoid (\%)}}
    & \cellcolor{white}\shortstack{\textbf{Driver}\\\textbf{Interventions}} \\
\midrule
\multicolumn{6}{@{}l}{\textbf{(a) Ego-Action Inputs}} \\
Without & $90.2$ & $\mathbf{40.8}$ & $\mathbf{91.5}$ & $\mathbf{48.0}$ & $\mathbf{8}$ \\
With & $\mathbf{90.7}$ & $39.1$ & $13.3$ & $3.4$ & $63$ \\
\midrule
\addlinespace[2pt]
\multicolumn{6}{@{}l}{\textbf{(b) Block-Wise Injection}} \\
All 28 Layers & $\mathbf{90.3}$ & $\mathbf{40.3}$ & $\mathbf{89.9}$ & $\mathbf{76.3}$ & $\mathbf{9}$ \\
Last Layer Only & $90.2$ & $39.8$ & $20.4$ & $0.0$ & $76$ \\
\midrule
\addlinespace[2pt]
\multicolumn{6}{@{}l}{\textbf{(c) Current-Frame Supervision}} \\
Without & $91.0$ & $44.0$ & -- & -- & $13$ \\
With & $\mathbf{91.3}$ & $\mathbf{44.5}$ & -- & -- & $\mathbf{8}$ \\
\bottomrule
\end{tabular*}}%
\kern0pt
\end{minipage}
\end{lrbox}

\noindent\usebox{\ablationInputTable}\hfill
\begin{minipage}[b][\ht\ablationInputTable][s]{0.47\linewidth}
    \raggedright
    \begingroup
        \caption{\textbf{TrajLoop and TrajAE.} On NAVSIM v1 navtest~\cite{bench1}, we add recurrent conditioning and then trajectory latents to the TrajDiT baseline.}
\label{tab:ablation-loop-ae}
\footnotesize
\setlength{\tabcolsep}{1.15pt}
\begin{tabular*}{\linewidth}{@{\extracolsep{\fill}}lccccc>{\columncolor{navsimScoreBlue!10}[\tabcolsep][0pt]}c@{}}
\toprule
\textbf{Module} & \textbf{NC} & \textbf{DAC} & \textbf{EP} & \textbf{TTC} & \textbf{C.} & \cellcolor{white}\shortstack{\textbf{PDMS}\\$\uparrow$} \\
\midrule
\textbf{TrajDiT (Base)}& $99.1$& $98.0$& $84.4$& $97.2$& $100.0$& $91.1$\\
\textbf{+TrajLoop}& $99.2$& $98.2$& $84.8$& $97.3$& $100.0$& $91.4$\\
\textbf{+TrajAE}&$\mathbf{99.3}$&$\mathbf{98.5}$&$\mathbf{85.0}$&$\mathbf{97.4}$&$\mathbf{100.0}$&$\mathbf{91.9}$\\
\bottomrule
\end{tabular*}

        \par
    \endgroup
    \vspace{7pt plus 1fill}
    \caption{\textbf{Long-horizon prediction.} We evaluate a fixed 1\,s horizon and two sampled windows, measuring compliance and driver interventions on the vehicle.}
\label{tab:ablation-horizon}
\footnotesize
\begingroup
\setlength{\tabcolsep}{1.6pt}
\par\nointerlineskip
\hbox{\begin{tabular*}{\linewidth}{@{\extracolsep{\fill}}lcc>{\columncolor{navsimScorePurple!10}[\tabcolsep][0pt]}c@{}}
\toprule
\textbf{Horizon}
& \shortstack{\textbf{Navigation}\\\textbf{(\%)}$\uparrow$}
& \shortstack{\textbf{Traffic Lights}\\\textbf{(\%)}$\uparrow$}
& \cellcolor{white}\shortstack{\textbf{Driver}\\\textbf{Interventions}$\downarrow$} \\
\midrule
\rule[-4.36pt]{0pt}{8pt}Fixed 1 s
& $86.0$ & $76.2$ & $23$ \\
\rule[-4.36pt]{0pt}{8pt}Sampled 1--10 s
& $94.0$ & $85.7$ & $9$ \\
\rule[-4.36pt]{0pt}{8pt}Sampled 1--20 s
& $\mathbf{96.0}$ & $\mathbf{95.2}$ & $\mathbf{5}$ \\
\bottomrule
\end{tabular*}}%
\endgroup

    \par\kern0pt
\end{minipage}
\end{table}

\noindent\textbf{Ego-action inputs.}
Panel (a) of Tab.~\ref{tab:ablation-input-supervision} shows the effect of providing the current ego action to the Qwen3-VL 2B backbone~\cite{bai2025qwen3vl}. Removing this input lowers NAVSIM v1 PDMS from 90.7 to 90.2 but raises navhard EPDMS from 39.1 to 40.8. On R1, start-up success rises from 13.3\% to 91.5\%, braking and avoidance success rises from 3.4\% to 48.0\%, and the intervention count falls from 63 to 8. During road testing, the variant conditioned on the current ego action tended to continue that action even as traffic conditions changed. Thus, \model~omits the current ego action as an input and learns to infer ego motion, including changes in speed, from successive camera observations.

\Needspace{5\baselineskip}
\noindent\textbf{Block-wise injection.}
Panel (b) of Tab.~\ref{tab:ablation-input-supervision} compares a planner that reads all 28 backbone layers with one that reads the final layer. They achieve comparable PDMS and EPDMS, but the layer-aligned planner raises start-up success from 20.4\% to 89.9\% and braking and avoidance success from 0.0\% to 76.3\% on R1. Interventions fall from 76 to 9. Access to intermediate features is associated with more responsive starts and braking on R1, a difference that is less apparent from the aggregate benchmark scores.

\noindent\textbf{Joint current and future supervision.}
EponaV2~\cite{xu2026eponav2} reports gains from future supervision over current-frame supervision alone. Panel (c) of Tab.~\ref{tab:ablation-input-supervision} tests their combination. Adding current-frame supervision raises v1 navtest PDMS from 91.0 to 91.3 and v2 navhard EPDMS from 44.0 to 44.5. On R1, interventions fall from 13 to 8. The current targets supervise observed geometry and semantics, and the future targets supervise how those cues evolve under the observed motion over the sampled prediction horizon.

\noindent\textbf{Long-horizon prediction.}
Tab.~\ref{tab:ablation-horizon} compares three prediction windows in real-vehicle tests. From a fixed 1\,s horizon to a sampled window of 1 to 20\,s, navigation compliance rises from 86.0\% to 96.0\%, traffic-light compliance rises from 76.2\% to 95.2\%, and interventions fall from 23 to 5. The intermediate window yields intermediate scores, with the largest compliance gain over the full range occurring at traffic lights.

\noindent\textbf{TrajLoop conditioning.}
Tab.~\ref{tab:ablation-loop-ae} shows that recurrent condition refinement improves the TrajDiT baseline from \textbf{91.1} to \textbf{91.4 PDMS} on NAVSIM v1 navtest with the same trajectory representation.

\noindent\textbf{Trajectory representation.}
Adding TrajAE to TrajLoop raises PDMS from \textbf{91.4} to \textbf{91.9} in Tab.~\ref{tab:ablation-loop-ae}. Fig.~\ref{fig:ablation_loop_trajae_qualitative} illustrates both ablation steps in six navtest scenes. The 12-scene review set contains cases where average displacement error decreased at both steps. We sorted scenes by token and used seed 2027 to sample six without replacement. The figure reports the errors in the order of the three variants in the legend.

\begin{figure}[!htp]
    \centering
    \begin{minipage}[t]{0.487\linewidth}
        \vspace{0pt}\centering
        \begin{minipage}[c][58pt][c]{\linewidth}
            \centering\includegraphics[width=\linewidth]{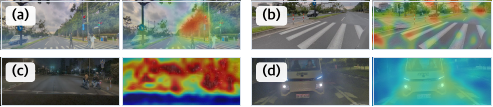}
        \end{minipage}
        \captionsetup{font=small}
        \caption{\textbf{Multi-view TrajAE alignment.} RGB images are paired with cosine overlays. Daytime panels (a) and (b) use per-view scaling. Panels (c) and (d) show the front and rear of one nighttime scene at the original scale.}
        \label{fig:trajae_cosine}
    \end{minipage}\hfill
    \begin{minipage}[t]{0.487\linewidth}
        \vspace{0pt}\centering
        \begin{minipage}[c][58pt][c]{\linewidth}
            \centering\includegraphics[width=\linewidth]{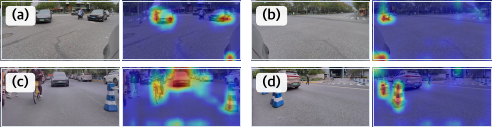}
        \end{minipage}
        \captionsetup{font=small}
        \caption{\textbf{Planner attention in real driving.} Each scene pairs an RGB image with its attention overlay. The overlays highlight vehicles, riders, and traffic cones. Fig.~\ref{fig:real_driving_attention_cases} extends the comparison across all six cameras.}
        \label{fig:driving_attention}
    \end{minipage}
\end{figure}

\noindent\textbf{Trajectory features and planner attention.}
To visualize TrajAE alignment, we compute cosine similarity between projected semantic trajectory tokens and final-layer visual tokens. Fig.~\ref{fig:trajae_cosine} shows stronger responses around parts of a crosswalk and nearby road users, including in side and rear cameras. Fig.~\ref{fig:multiview_cosine_cases} shows similar responses across all six views in daytime and nighttime road scenes.

\FloatBarrier

Fig.~\ref{fig:driving_attention} shows attention near vehicles, riders, and traffic cones. We average attention over the five flow steps and all 28 layers. The six-camera examples in Fig.~\ref{fig:real_driving_attention_cases} show these responses across daytime and nighttime road scenes with cross traffic, pedestrians, road barriers, and scooters.

\FloatBarrier

\section{Conclusion}
\label{sec:conclusion}



We presented \model, a world-action model that combines compact multi-view inputs, structured navigation, and long-horizon prediction in a shared temporal backbone. TrajLoop conditions trajectory refinement on features from all backbone layers, while prediction heads supervise training without adding scene generation at deployment. With approximately 35K hours of driving data, the policy achieves strong single-sample NAVSIM results and sustains 10-Hz onboard planning over approximately 240~km of road testing across four urban routes in Shanghai, China. On route R1, using all backbone layers reduced driver interventions from 76 to 9 relative to using the last layer alone, despite similar NAVSIM scores. This result illustrates why we evaluate planning both on benchmarks and on the vehicle.

\Needspace{23\baselineskip}
\section{Author Contributions}
\label{sec:authors}

\begingroup
\newcommand{\eponaxRoleMark}[1]{%
  {\normalfont\fontsize{8pt}{8pt}\selectfont\textcolor{reportDeepAccent}{\ensuremath{#1}}}}
\newcommand{\eponaxAuthor}[2][]{%
  \mbox{\textbf{#2}\ifstrempty{#1}{}{\hspace{0.1em}\raisebox{0.55ex}{\eponaxRoleMark{#1}}}}}
\newcommand{\contributionheading}[1]{%
  \par\addvspace{4pt}\noindent\reporthead{#1}\par}

\begin{tcolorbox}[eponaxPanel]
  \small\setstretch{1}
  \setlength{\parskip}{0pt}
  \renewcommand{\arraystretch}{1.15}
  \noindent\begin{tabularx}{\linewidth}{@{}*{5}{>{\raggedright\arraybackslash}X}@{}}
    \eponaxAuthor[\ast\dagger]{Mingxiao Li} &
    \eponaxAuthor[\ast\dagger]{Zhijian Shu} &
    \eponaxAuthor[\ast\dagger]{Wei Yin} &
    \eponaxAuthor[\ast\dagger]{Jiawei Xu} &
    \eponaxAuthor[\ast\dagger]{Mingkai Jia} \\
    \eponaxAuthor[\ast\dagger]{Alan Liang} &
    \eponaxAuthor[\ast\dagger]{Zhizhou Zhong} &
    \eponaxAuthor[\ast]{Shiyao Li} &
    \eponaxAuthor[\ast]{Dongyue Lu} &
    \eponaxAuthor{Liang Yue} \\
    \eponaxAuthor{Zeyuan Xu} &
    \eponaxAuthor{Kaicheng Zhang} &
    \eponaxAuthor{Xiao-Xiao Long} &
    \eponaxAuthor{Lingdong Kong} &
    \eponaxAuthor[\ddagger]{Qian Zhang}
  \end{tabularx}\par
  \vspace{3pt}
  \noindent{\footnotesize\sffamily\color{reportMuted}
    \makebox[0.335\linewidth][l]{\eponaxRoleMark{\ast}\enspace Core contributor}%
    \makebox[0.335\linewidth][l]{\eponaxRoleMark{\dagger}\enspace Co-first author}%
    \eponaxRoleMark{\ddagger}\enspace Project leader\par}
  \tcblower
  \small\setstretch{1}
  \setlength{\parskip}{1pt}
  \renewcommand{\arraystretch}{1.1}
  \noindent
  \begin{minipage}[t]{0.477\linewidth}
    \raggedright
    \noindent\reporthead{Data}\par
    Mingxiao~Li, Wei~Yin, Alan~Liang

    \contributionheading{Training}
    \noindent\begin{tabularx}{\linewidth}{@{}
      >{\raggedright\arraybackslash}p{53pt}
      >{\raggedright\arraybackslash}X@{}}
      \textbf{Pretraining} & Mingxiao~Li, Alan~Liang, Wei~Yin \\
      \textbf{Fine-Tuning} & Zhijian~Shu, Jiawei~Xu \\
      \textbf{RL} & Zhijian~Shu, Jiawei~Xu, Zhizhou~Zhong, Alan~Liang
    \end{tabularx}

    \contributionheading{Evaluation}
    Zhijian~Shu, Jiawei~Xu, Alan~Liang, Mingxiao~Li
  \end{minipage}\hfill
  \begin{minipage}[t]{0.477\linewidth}
    \raggedright
    \noindent\reporthead{Development}\par
    \noindent\begin{tabularx}{\linewidth}{@{}
      >{\raggedright\arraybackslash}p{87pt}
      >{\raggedright\arraybackslash}X@{}}
      \textbf{System Architecture \& Codebase} & Mingxiao~Li \\
      \textbf{DINO-Tok} & Mingkai~Jia, Mingxiao~Li, Zhijian~Shu \\
      \textbf{TrajAE} & Mingkai~Jia, Mingxiao~Li \\
      \textbf{MVTok} & Alan~Liang, Dongyue~Lu \\
      \textbf{Navigation Encoder} & Zhizhou~Zhong, Mingxiao~Li, Liang~Yue \\
       \textbf{TrajLoop} & Mingxiao~Li\\
    \end{tabularx}
  \end{minipage}

  \eponaxPanelNextRow
  \begin{minipage}[t]{0.477\linewidth}
    \raggedright
    \noindent\reporthead{Writing}\par
    Zhijian~Shu, Lingdong~Kong, Alan~Liang, Mingxiao~Li, Mingkai~Jia,
    Zhizhou~Zhong, Jiawei~Xu, Shiyao~Li, Zeyuan~Xu
  \end{minipage}\hfill
  \begin{minipage}[t]{0.477\linewidth}
    \raggedright
    \noindent\reporthead{Project Advisor}\par
    Wei~Yin

    \contributionheading{Project Leadership}
    Qian~Zhang
  \end{minipage}
\end{tcolorbox}
\endgroup

\clearpage
\appendix
\setcounter{figure}{0}
\setcounter{table}{0}
\setcounter{equation}{0}
\renewcommand{\thefigure}{S\arabic{figure}}
\renewcommand{\thetable}{S\arabic{table}}
\renewcommand{\theHfigure}{supp.\arabic{figure}}
\renewcommand{\theHtable}{supp.\arabic{table}}
\renewcommand{\theHequation}{supp.\arabic{equation}}
\section*{Supplementary Material}
\begin{figure}[!htp]
    \centering
    \includegraphics[page=1,height=0.66\textheight,width=\linewidth,keepaspectratio]{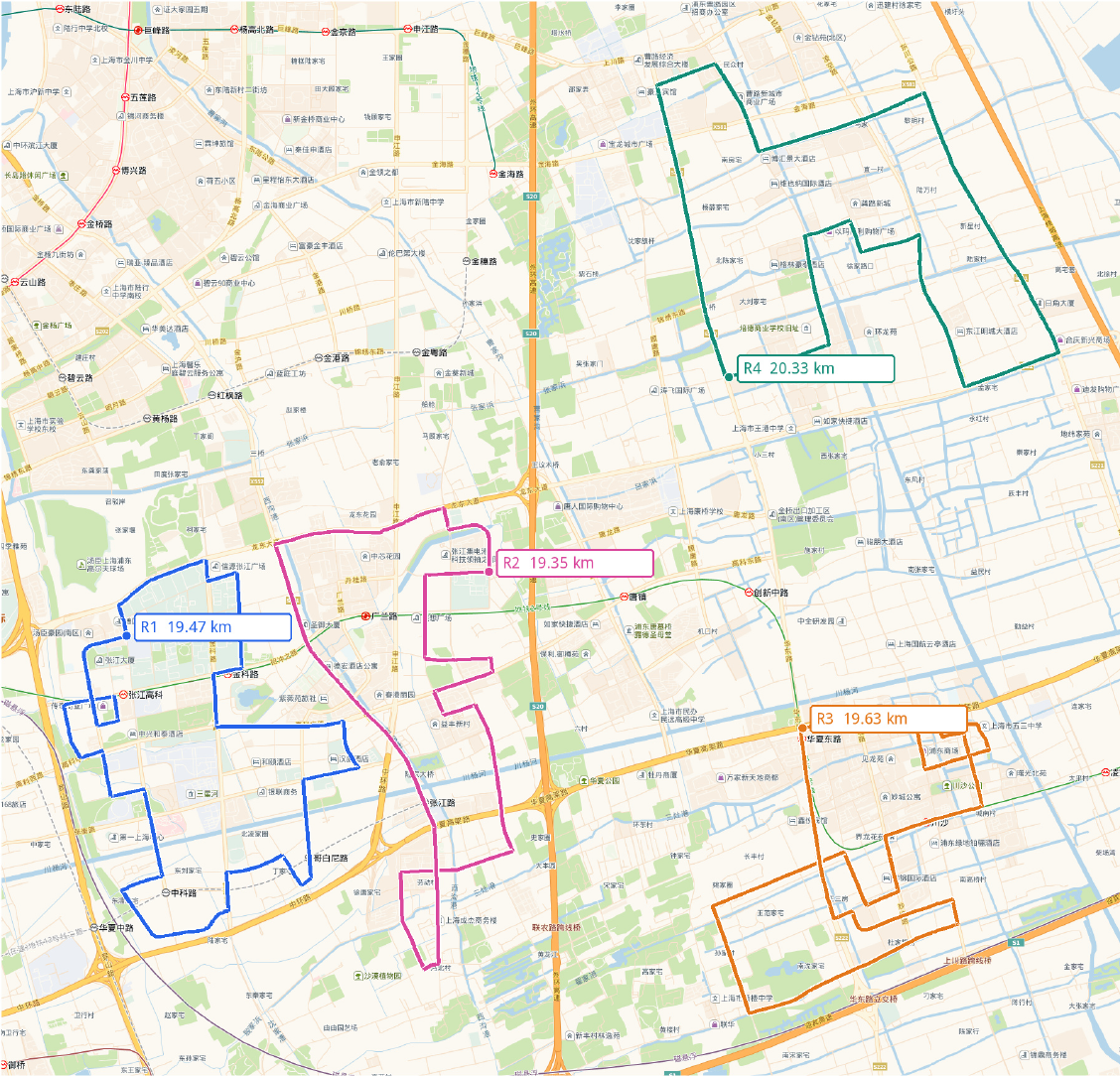}
    \caption{\textbf{Real-world evaluation routes in Shanghai, China.}
    Tab.~\ref{tab:route-locations} lists the route areas, estimated loop lengths, and selected streets. Map imagery is from Gaode Maps. Road geometry is credited to OpenStreetMap contributors under the Open Database License.}
    \label{fig:roads_appendix}
\end{figure}

\clearpage
\begin{figure}[!t]
    \ContinuedFloat
    \centering
    \begin{subfigure}[t]{0.48\linewidth}
        \centering
        \begin{minipage}[c][0.400\textheight][c]{\linewidth}
            \centering
            \includegraphics[page=2,width=\linewidth,height=0.400\textheight,keepaspectratio]{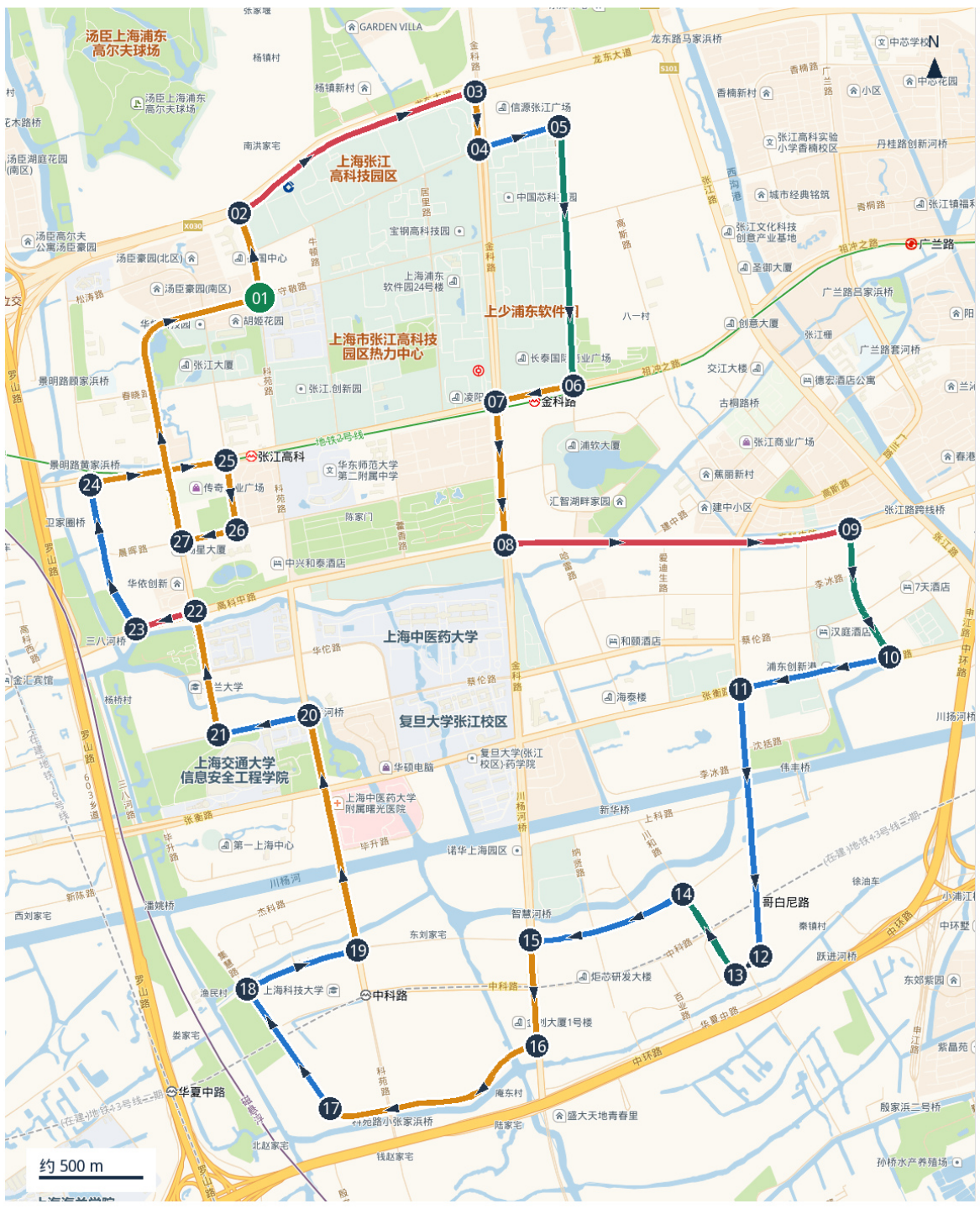}
        \end{minipage}
        \caption{R1. Zhangjiang Innovation Park.}
    \end{subfigure}\hfill
    \begin{subfigure}[t]{0.48\linewidth}
        \centering
        \begin{minipage}[c][0.400\textheight][c]{\linewidth}
            \centering
            \includegraphics[page=3,width=\linewidth,height=0.400\textheight,keepaspectratio]{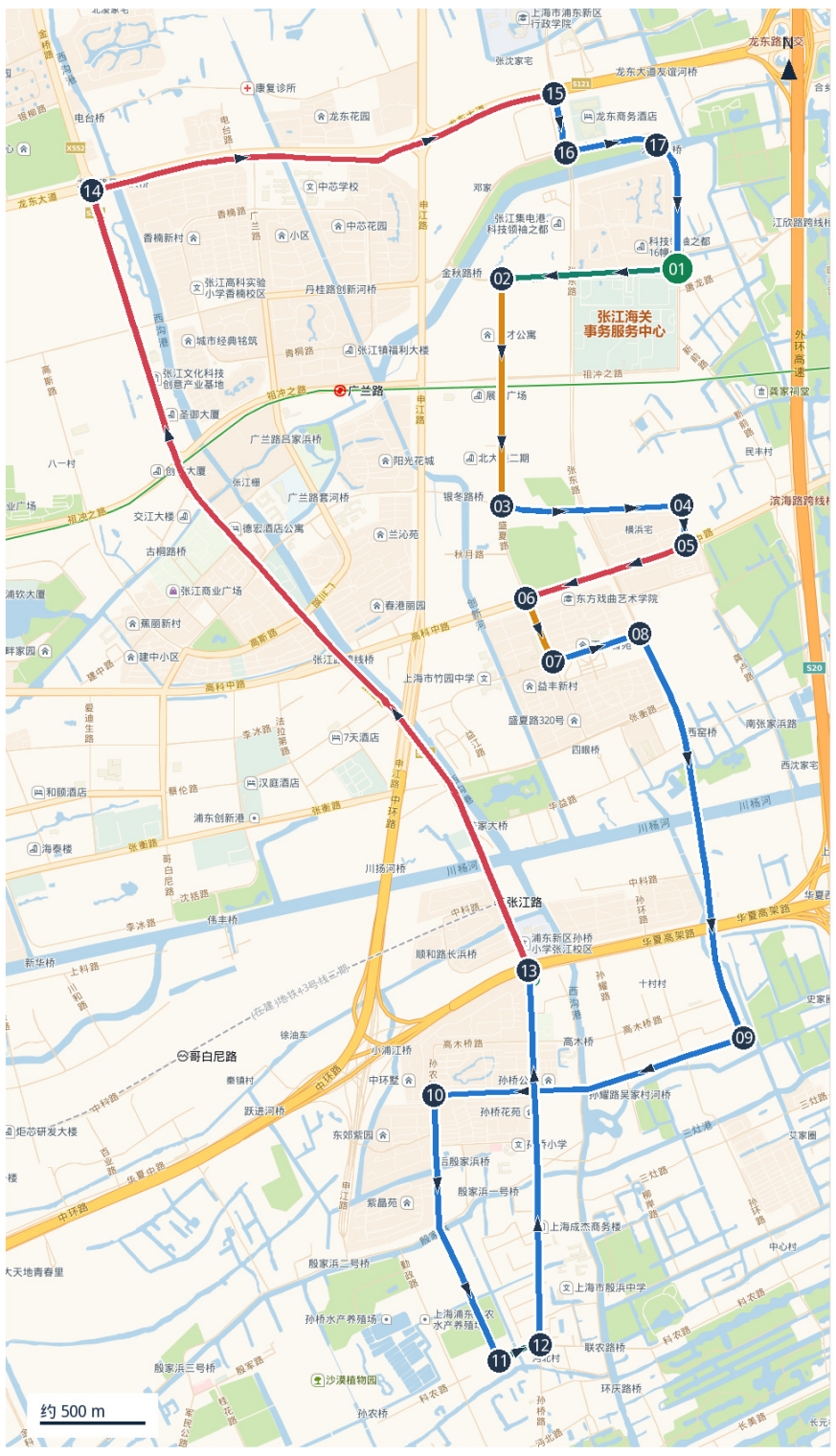}
        \end{minipage}
        \caption{R2. Zhangjiang East and Sunqiao.}
    \end{subfigure}

    \smallskip
    \begin{subfigure}[t]{0.48\linewidth}
        \centering
        \begin{minipage}[c][0.400\textheight][c]{\linewidth}
            \centering
            \includegraphics[page=4,width=\linewidth,height=0.400\textheight,keepaspectratio]{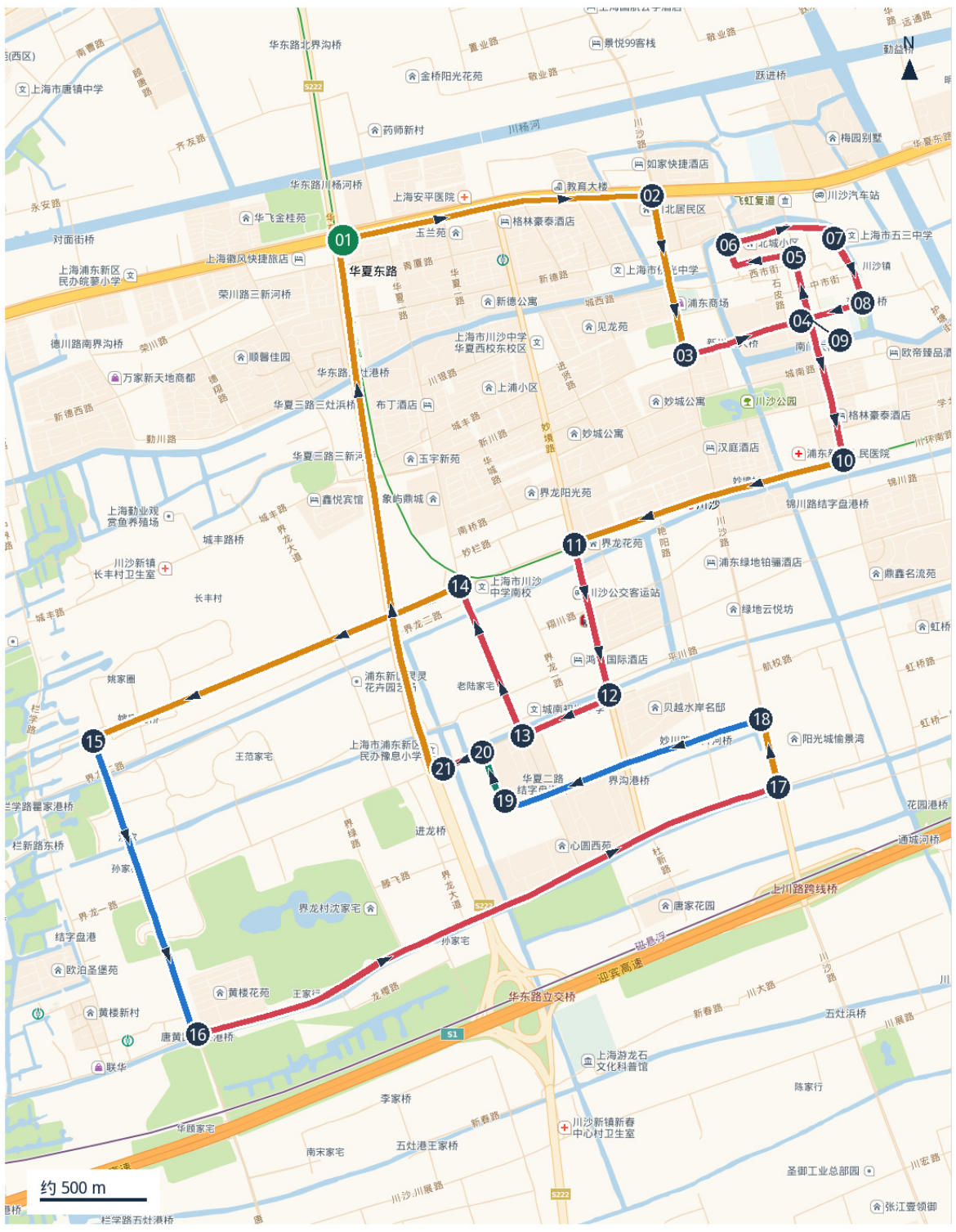}
        \end{minipage}
        \caption{R3. Chuansha via Chuanzhou Hwy.}
    \end{subfigure}\hfill
    \begin{subfigure}[t]{0.48\linewidth}
        \centering
        \begin{minipage}[c][0.400\textheight][c]{\linewidth}
            \centering
            \includegraphics[page=5,width=\linewidth,height=0.400\textheight,keepaspectratio]{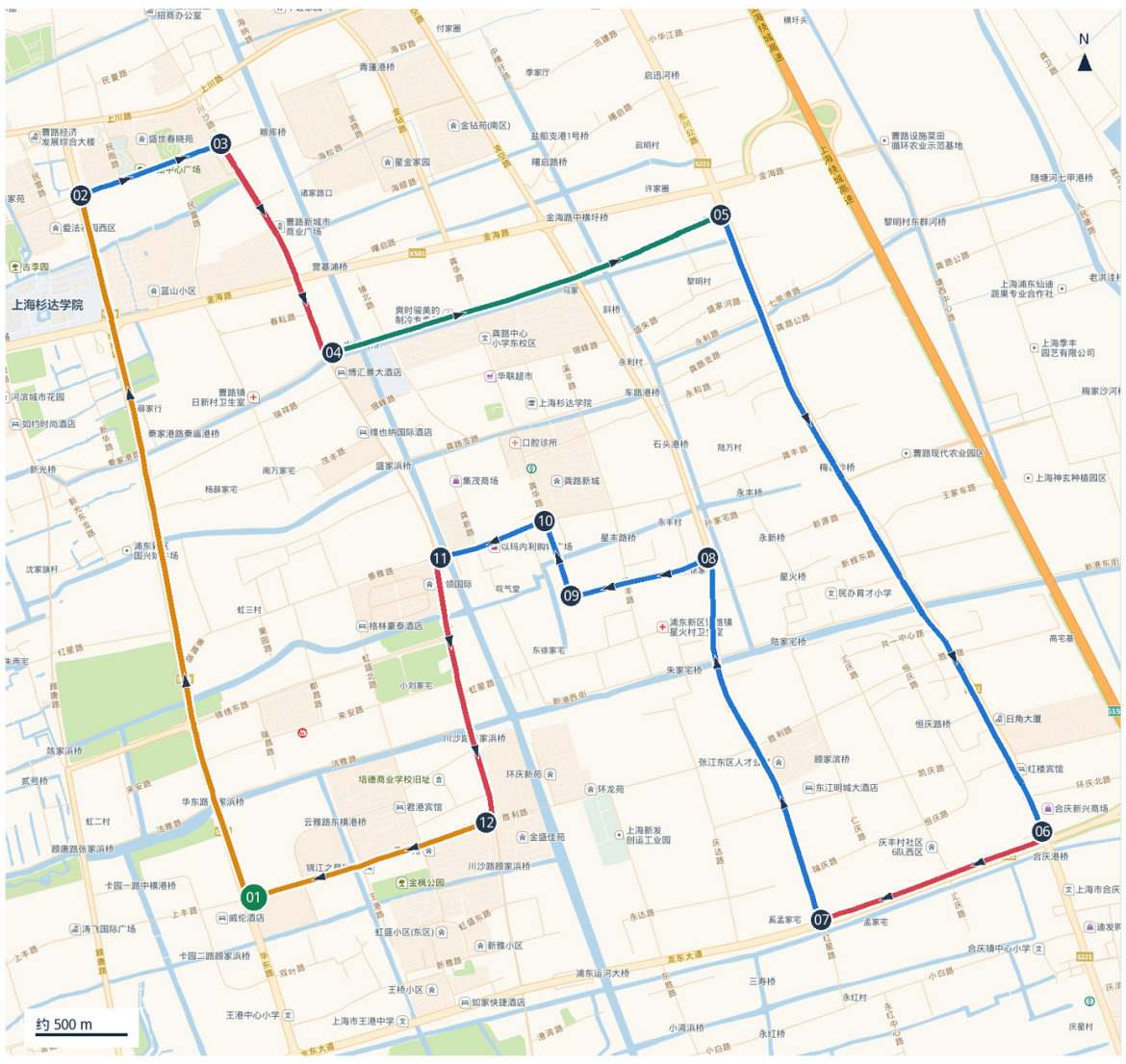}
        \end{minipage}
        \caption{R4. Caotanghe via Dongchuan Hwy.}
    \end{subfigure}
    \caption[]{\textbf{Detailed maps of the Shanghai test loops.}
        Numbered markers indicate intersections. Green, blue, orange, and red segments denote road classes I to IV.
        Basemap from Gaode Maps. Road geometry from OpenStreetMap contributors under the Open Database License.}
\end{figure}

\clearpage
\begin{figure}[!htbp]
    \centering
    \includegraphics[page=1,width=\linewidth]{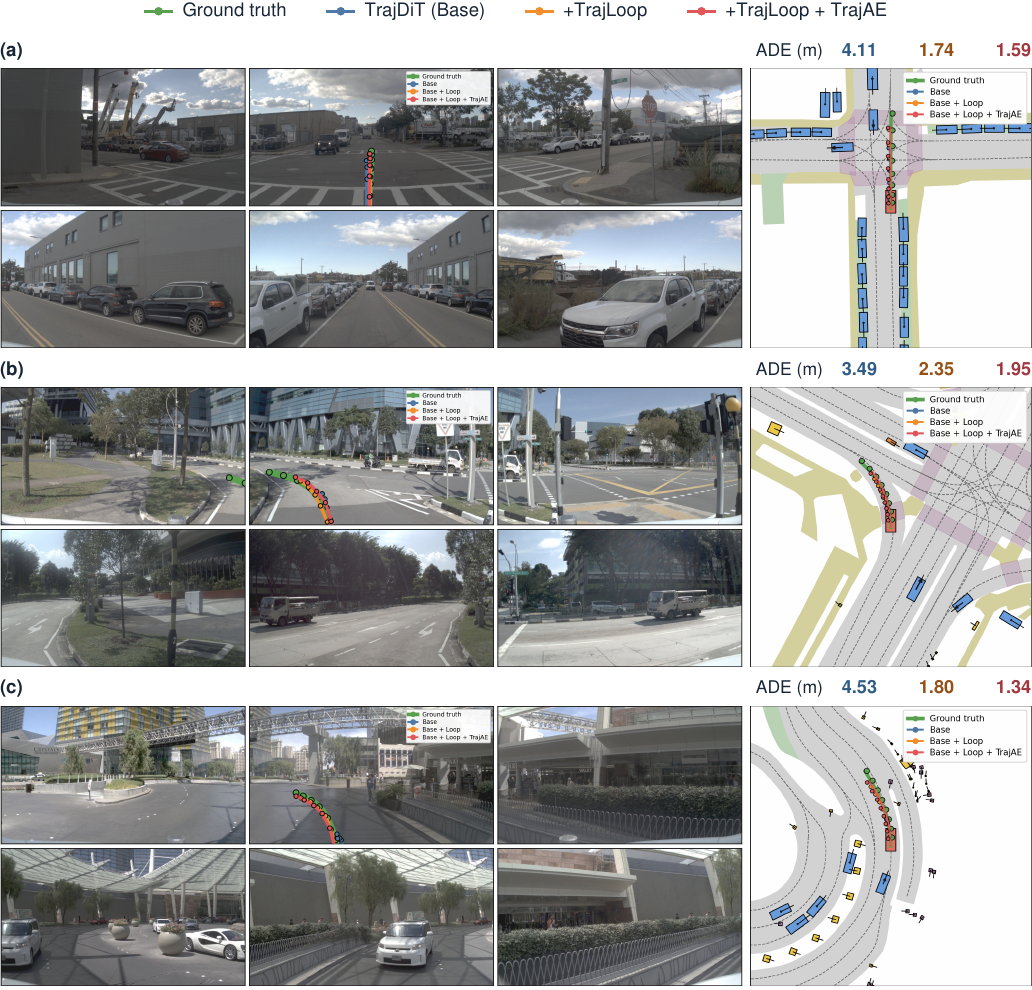}
    \caption{\textbf{Qualitative ablation of TrajLoop and TrajAE on NAVSIM~\cite{bench1} navtest.}
    Each panel shows trajectory overlays in six camera views and the corresponding bird's-eye view. The average displacement errors follow the three model variants in legend order. Tab.~\ref{tab:ablation-loop-ae} reports PDMS and its component metrics for the same three variants across the NAVSIM v1 navtest split.}
    \label{fig:ablation_loop_trajae_qualitative}
\end{figure}

\clearpage
\begin{figure}[!t]
    \ContinuedFloat
    \centering
    \includegraphics[page=2,width=\linewidth]{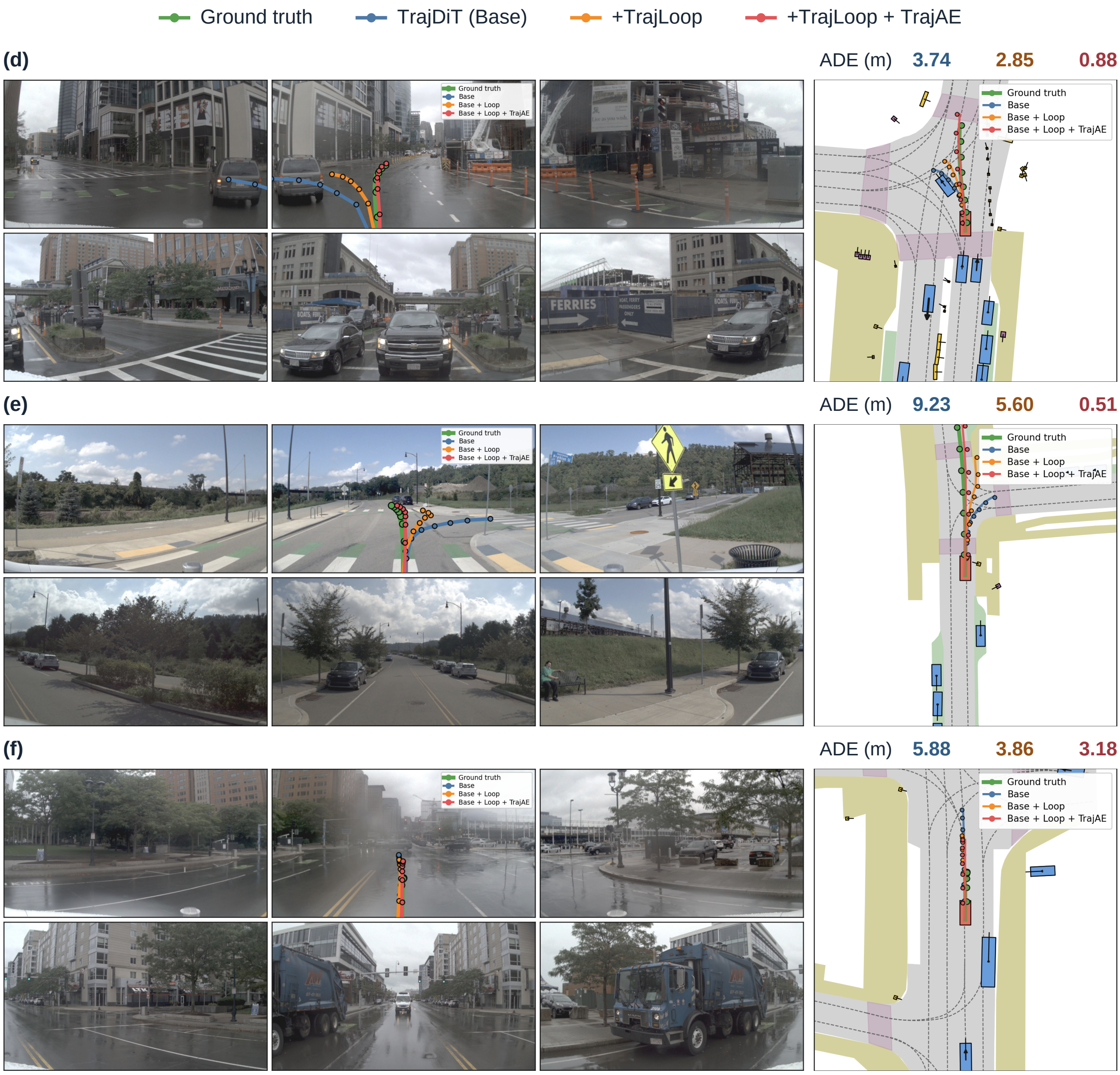}
    \caption[]{\textbf{Qualitative ablation of TrajLoop and TrajAE on NAVSIM navtest (continued).}
    Panels (d)--(f) use the same camera layout, trajectory legend, and ADE order as panels (a)--(c).}
\end{figure}

\clearpage
\begin{figure}[!p]
    \centering
    \includegraphics[width=\linewidth,height=0.87\textheight,keepaspectratio]{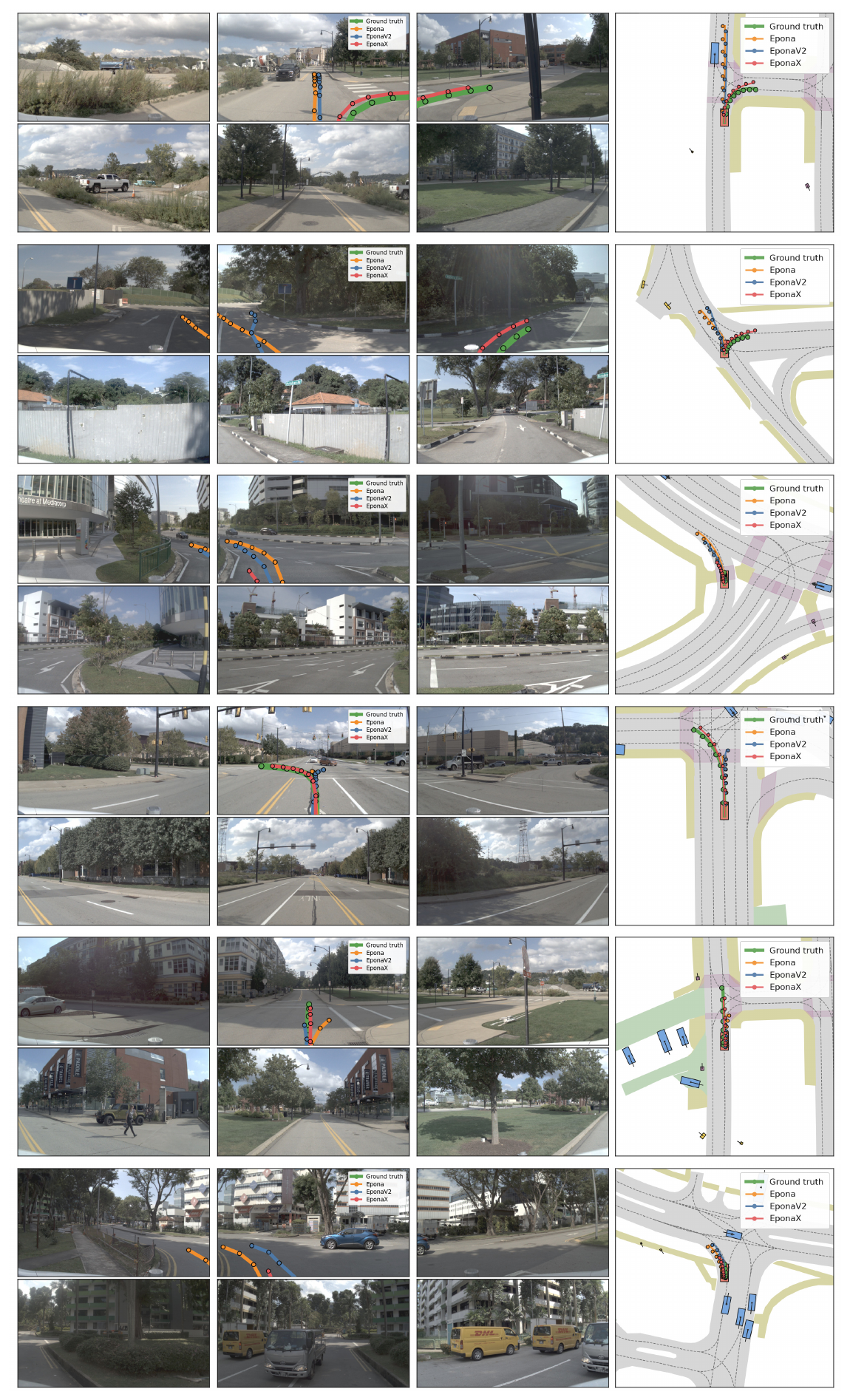}
    \caption{\textbf{Trajectory comparisons across the Epona series.} Each row pairs the camera views with a bird's-eye view of the same example. Green denotes the recorded trajectory, and orange, blue, and red denote Epona, EponaV2, and \model, respectively. The selected examples illustrate differences in turn direction, curvature, and progress against the recorded trajectory.}
    \label{fig:series_trajectory_cases}
\end{figure}

\begin{figure}[!p]
    \centering
    \includegraphics[width=\linewidth,height=0.87\textheight,keepaspectratio]{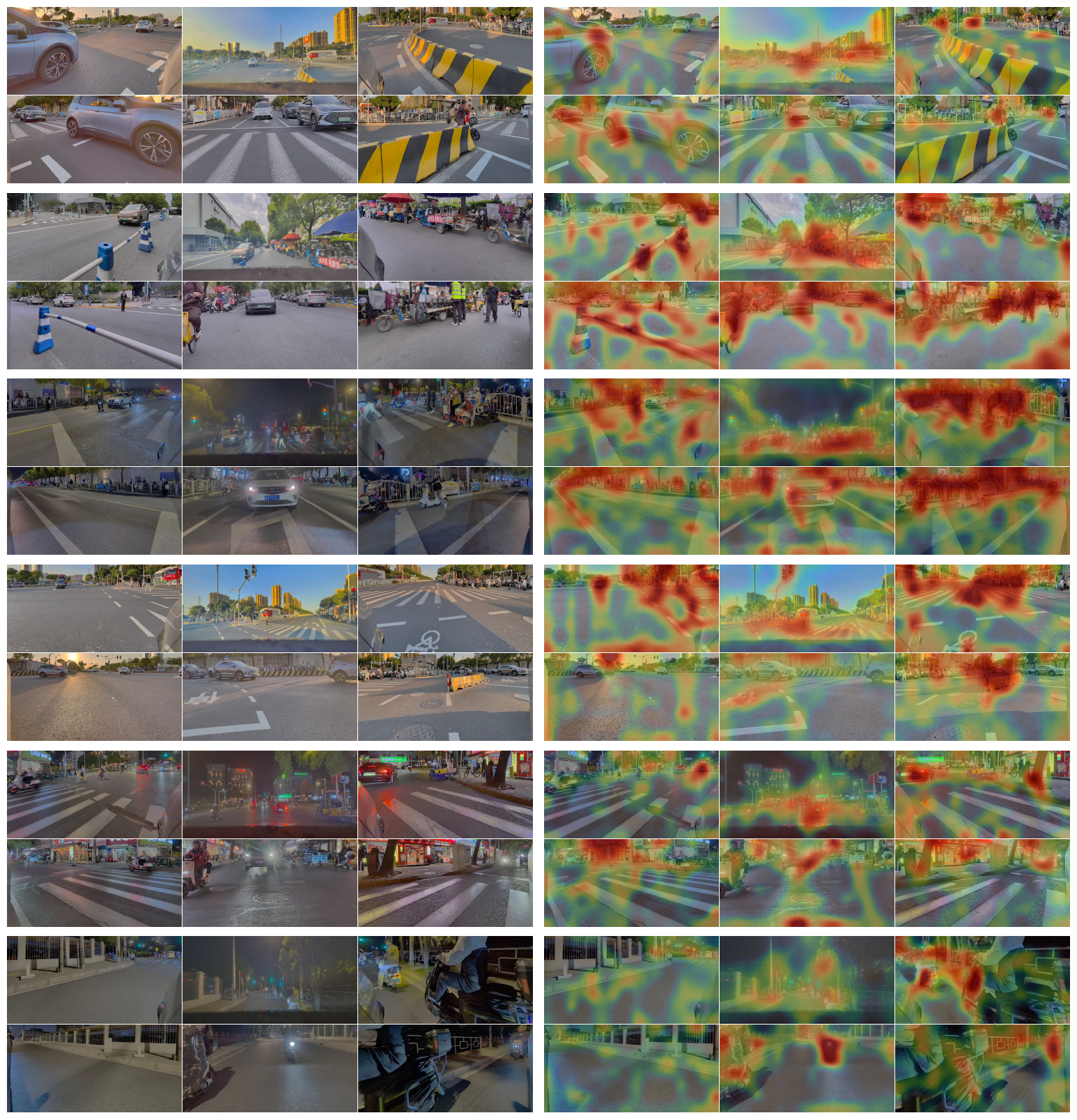}
    \caption{\textbf{Multi-view TrajAE cosine similarity in additional real driving scenes.} Each scene groups six camera inputs (left) and the corresponding overlays (right). The overlays show cosine similarity between projected trajectory semantic tokens and world-model visual features. Responses appear around vehicles, riders, and parts of the road in both daytime and nighttime views.}
    \label{fig:multiview_cosine_cases}
\end{figure}

\begin{figure}[!p]
    \centering
    \includegraphics[width=\linewidth,height=0.87\textheight,keepaspectratio]{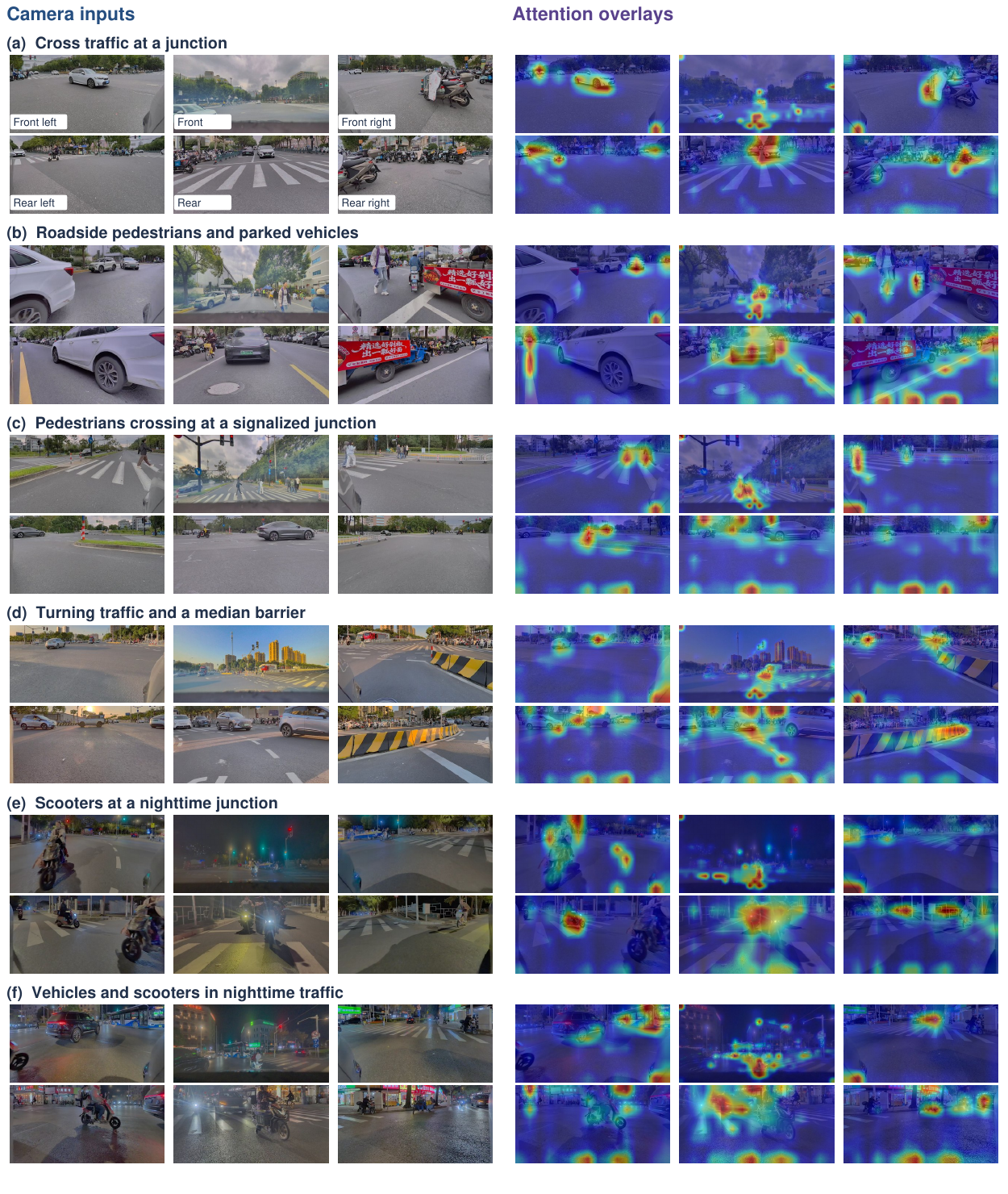}
    \caption{\textbf{Attention across six cameras in real driving.} Each case shows the camera inputs (left) and the corresponding attention overlays (right). The source visualizations average attention over five denoising steps and 28 layers. The selected scenes include cross traffic, pedestrians, road barriers, and scooters across daytime and nighttime driving conditions.}
    \label{fig:real_driving_attention_cases}
\end{figure}

\clearpage
\phantomsection
\addcontentsline{toc}{section}{References}
{\setlength{\bibsep}{1.5pt}
 \bibliographystyle{vitalplainnat}
 \bibliography{main}}
\end{document}